\documentclass[]{amap}

\usepackage[toc,page,header]{appendix}
\usepackage{minitoc}
\usepackage{solarized-light}
\usepackage{booktabs}
\usepackage{multirow}
\usepackage{graphicx}
\usepackage{array}
\usepackage{siunitx,array}
\usepackage[table]{xcolor}
\usepackage{makecell,array,booktabs}
\usepackage{makecell}
\usepackage{framed}

\usepackage{bbding}
\usepackage{hyperref}
\usepackage{amsmath}
\usepackage{amsfonts}
\usepackage{placeins}
\usepackage[colorinlistoftodos]{todonotes}
\usepackage{longtable}
\usepackage{hhline}
\usepackage{fancyvrb}
\usepackage{graphicx}
\usepackage[table]{xcolor}
\usepackage{booktabs}
\usepackage{float}
\usepackage{fvextra}
\usepackage{CJKutf8}
\usepackage{multicol}
\usepackage{float}
\usepackage{placeins}
\usepackage{cleveref}
\usepackage{tablefootnote}
\usepackage{threeparttable}
\usepackage{tabularx}
\usepackage{mdframed}
\usepackage{subcaption}
\usepackage[usestackEOL]{stackengine}
\usepackage[numbers]{natbib}
\newcommand{\commentout}[1]{}
\renewcommand{\paragraph}[1]{\noindent\textbf{#1.}\hspace*{1em}}
\usepackage{enumitem}
\usepackage{hyperref}
\setlist[itemize]{leftmargin=15pt}
\usepackage{siunitx,array}
\usepackage{tabularx}
\usepackage{adjustbox}
\usepackage{arydshln}
\usepackage{pifont}
\usepackage{bbding}
\definecolor{ampblue}{rgb}{0.82, 0.88, 0.94}
\usepackage[table]{xcolor} % 用于颜色
\usepackage{array} 
\usepackage[dvipsnames]{xcolor}

\RequirePackage{xspace}
\makeatletter
\DeclareRobustCommand\onedot{\futurelet\@let@token\@onedot}
\def\@onedot{\ifx\@let@token.\else.\null\fi\xspace}

\makeatother
\newcommand{\VLAFM}{ABot-M0}

\usepackage{xcolor}

\definecolor{abot1}{HTML}{0185FE}
\definecolor{abot2}{HTML}{0185FE}
\definecolor{abot3}{HTML}{0185FE}
\definecolor{abot4}{HTML}{0185FE}
\definecolor{abot5}{HTML}{FB8C00}
\definecolor{abot6}{HTML}{FB8C00}
\definecolor{abot7}{HTML}{FB8C00}

\newcommand{\ABotMZero}{%
{\color{abot1}A}%
{\color{abot2}B}%
{\color{abot3}o}%
{\color{abot4}t}%
{\color{abot5}-}%
{\color{abot6}M}%
{\color{abot7}0}%
}

\title{\ABotMZero: VLA Foundation Model for Robotic Manipulation with Action Manifold Learning}

\author{AMAP CV Lab}
\vspace{-11pt}

\contribution{See \hyperref[sec:contributions]{Contributions} section for a full author list.}

\abstract{

Building general-purpose embodied agents across diverse hardware remains a central challenge in robotics, often framed as the ``one-brain, many-forms'' paradigm. Progress is hindered by fragmented data, inconsistent representations, and misaligned training objectives.
We present \textbf{\VLAFM{}}, a framework that builds a systematic data curation pipeline while jointly optimizing model architecture and training strategies, enabling end-to-end transformation of heterogeneous raw data into unified, efficient representations. From six public datasets, we clean, standardize, and balance samples to construct \textbf{UniACT-dataset}, a large-scale dataset with over 6 million trajectories and 9,500 hours of data, covering diverse robot morphologies and task scenarios. Unified pre-training improves knowledge transfer and generalization across platforms and tasks, supporting general-purpose embodied intelligence.
To improve action prediction efficiency and stability, we propose the \textit{Action Manifold Hypothesis}: effective robot actions lie not in the full high-dimensional space but on a low-dimensional, smooth manifold governed by physical laws and task constraints. Based on this, we introduce \textbf{Action Manifold Learning (AML)}, which uses a DiT backbone to predict clean, continuous action sequences directly. This shifts learning from denoising to projection onto feasible manifolds, improving decoding speed and policy stability.
\VLAFM{} supports modular perception via a dual-stream mechanism that integrates VLM semantics with geometric priors and multi-view inputs from plug-and-play 3D modules such as VGGT and Qwen-Image-Edit, enhancing spatial understanding without modifying the backbone and mitigating standard VLM limitations in 3D reasoning.
Experiments show components operate independently with additive benefits. We will release all code and pipelines for reproducibility and future research.

\bigskip

\textbf{Date:} February 11, 2026

\textbf{Code:} \url{https://github.com/amap-cvlab/ABot-Manipulation}

\textbf{Project Page:} \url{https://amap-cvlab.github.io/ABot-Manipulation}

}

\begin{document}
\maketitle
\vspace{-4pt}

\begin{figure}[!h]
    \centering
    \vspace{-10pt}
\includegraphics[width=0.6\linewidth]{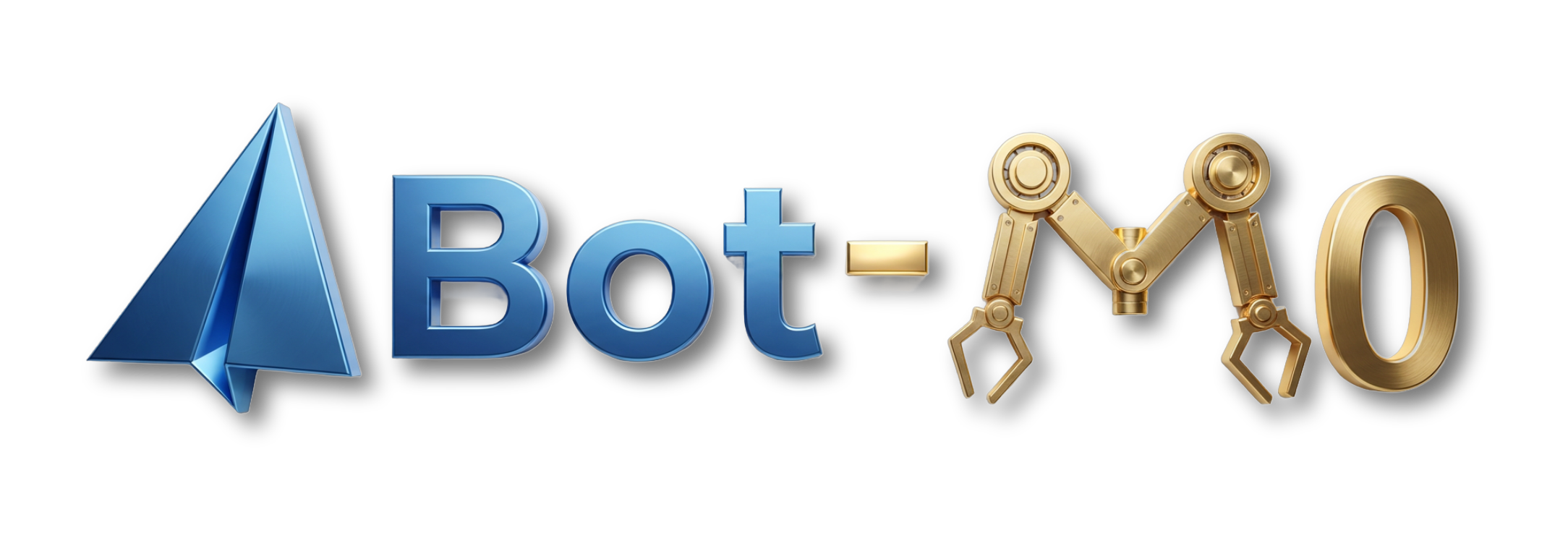}
    % \caption{\textbf{xxx}.}
    \label{fig:model}
\end{figure} 

% \begin{figure}[!h]
%     \centering
%     \vspace{-10pt}
% \includegraphics[width=1.0\linewidth]{figure/teaser1.png}
%     \caption{\textbf{\VLAFM{}}.}
%     \label{fig:model}
% \end{figure}

% \begin{figure}[!h]
%     \centering
%     \vspace{-10pt}
% \includegraphics[width=1.0\linewidth]{figure/teaser2.png}
%     \caption{\textbf{\VLAFM{}}.}
%     \label{fig:model}
% \end{figure}

% \begin{figure}[!h]
%     \centering
%     \vspace{-10pt}
% \includegraphics[width=1.0\linewidth]{figure/白色背景.png}
%     \caption{\textbf{\VLAFM{}}.}
%     \label{fig:model}
% \end{figure}

\newpage
\tableofcontents
\newpage

%\newpage
\section{Introduction}
\label{sec:intro}

The ultimate goal of robotics is to build general embodied agents that operate across diverse robotic bodies, realizing the ``one brain, many forms'' vision~\cite{rt2,liu2026rdt2,cheang2025gr}. Such an agent must perceive its environment, understand natural language instructions, reason about actions, and learn from interaction~\cite{rt2,liu2026rdt2,cheang2025gr,zheng25}. While Large Language Models (LLMs) have achieved universality in the digital realm, embodied intelligence remains fragmented. Despite decades of progress, this vision remains unrealized. Policies trained on one robot rarely transfer to another, and high-level semantic reasoning struggles to translate into low-level physical precision. Embodied intelligence has yet to reach its transformative moment.

We argue that the field currently faces three fundamental barriers. First, data scale remains below critical mass. Unlike standard vision-language data, embodied data require precise action labels, which are expensive and slow to collect. Most current approaches rely on pre-training sets limited to a single robot type or platform. While effective for narrow deployment, this restricts the development of general policies by limiting exposure to diverse morphologies and task distributions. Second, data quality varies widely and lacks standardization. Action representations, coordinate systems, and control frequencies differ across datasets, making it difficult for models to extract common patterns across embodiments. The resulting fragmentation introduces noise and weakens learning, as models must allocate capacity to memorize idiosyncrasies rather than acquire transferable skills. Third, existing pre-training paradigms are mismatched. Most Vision-Language-Action (VLA) models start from Vision-Language Models (VLMs), whose visual encoders focus on semantic recognition rather than 3D structure or physical dynamics. Yet embodied behavior demands fine spatial understanding and causal reasoning, capabilities that cannot be acquired from 2D supervision alone, nor easily added through mechanisms such as Chain-of-Thought, because such methods operate at the reasoning level without enriching the underlying perceptual representation. Achieving general embodied intelligence therefore requires systematic improvements in data, representation, and training design.

We address a practical question: given the high cost and hardware dependence of robotic data collection, can we integrate global open-source datasets, currently isolated in silos, into a unified foundation for training VLA models? This points to a different direction: embodied intelligence need not emerge from closed, proprietary systems, but can instead develop through the aggregation of heterogeneous data and incremental capability building, with shared infrastructure enabling cumulative progress across research groups.

To this end, we introduce \VLAFM{} (\textbf{\uline{A}}map VLA Foundation Model for Ro\textbf{\uline{bot}}ic \textbf{\uline{M}}anipulation), a unified approach that jointly optimizes data processing and model architecture. We first construct the \textbf{UniACT-dataset}, combining six major open-source datasets to form a mixed training set of over 6 million trajectories spanning 9500+ hours and 20+ embodiments, which is currently the largest collection within the non-private domain. To resolve inconsistencies in action format, coordinate system, and sampling rate, we define a standardized preprocessing pipeline. All actions are converted to delta actions in the end-effector frame; rotations are encoded using rotation vectors to avoid singularities; and we apply a pad-to-dual strategy to support both single-arm and dual-arm tasks within a shared framework. During training, we conduct uniform sampling across datasets to balance the distribution of tasks and embodiments. This unified data foundation breaks down barriers between datasets and enables robust cross-embodiment generalization by aligning both geometric and temporal structures across sources.

Building on this unified data foundation, we construct the \textbf{\VLAFM} model based on a VLM followed by an action expert. We employ a two-stream feature interaction to enhance both semantic and 3D perception for action prediction and propose the Action Manifold Learning (AML) to improve the efficiency and stability of action prediction. First, most VLA models~\cite{GR00T,team2024octo,li2024cogact,reuss2025flower} train the policy to predict noise, either $\epsilon$-prediction or $v$-prediction, which is unstructured, meaningless, and often includes invalid actions such as jittering or discontinuities. These noisy targets spread across high-dimensional space, requiring large model capacity and leading to inefficient learning. Inspired by JiT~\cite{JIT} compilation, we propose the \textit{"Action Manifold Hypothesis"}: successful actions do not scatter randomly but lie on a low-dimensional, smooth manifold shaped by physics, task goals, and environmental constraints. Based on this, we design the Action Manifold Learning (AML) mechanism, where the DiT backbone directly predicts clean action sequences. This shifts the learning objective from fitting noise to projecting onto feasible action manifolds, improving decoding speed and policy stability by reducing output uncertainty. Second, to improve feature interaction between action expert and VLM, we design a dual-stream feature interaction. The action expert takes the feature of VLM's last layer to acquire visual and semantic reasoning. Additionally, a perception module injects geometry-aware features via optional models, such as VGGT~\cite{wang2025vggt} to compute scene-level structural features and Qwen-Image-Edit~\cite{wu2025qwenimagetechnicalreport} to fuse features from multi-view observations. The perception module acts as a plug-and-play expert that compensates for the perceptual limitations of standard vision-language models.

We perform full ablation studies under a consistent training setup, showing that data standardization, architectural changes, and training redesign are orthogonal and their benefits accumulate. Our model achieves average success rates of 98.6\%, 80.5\%, 58.3\% and 81.2\% on benchmarks including LIBERO~\cite{liu2023libero}, LIBERO-Plus~\cite{fei25libero-plus}, RoboCasa GR1 Tabletop Tasks~\cite{GR00T} and Robotwin2.0~\cite{chen2025robotwin}, outperforming strong baselines such as $\pi_{0.5}$~\cite{pi_0.5}, UniVLA~\cite{bu2025univla}, and OpenVLA-OFT~\cite{kim2024openvla}. These results validate a complete pipeline from data curation to architecture design to capability emergence, demonstrating that high-performance, generalizable embodied intelligence can be achieved through systematic engineering without reliance on proprietary data. The gains from each component prove consistent across platforms, indicating that learned representations transfer beyond dataset-specific biases. We will release our full data processing and training codebase to support open, collaborative progress toward general-purpose robotics.
\section{Dataset}  
\label{sec:dataset}  

To support large-scale pre-training for general embodied intelligence, we  conducted a comprehensive analysis of prevalent robotics datasets in \Cref{subsec:data_analysis}. Then we undertake two systematic data-level initiatives. First, in \Cref{subsec:data_process} we perform comprehensive data cleaning and quality governance to address prevalent issues in multi-source datasets, including format inconsistencies, missing instructions, and semantic noise. And in \Cref{subsec:data_format} we establish a unified data representation to enable cross-platform and cross-embodiment alignment of manipulation semantics. These efforts together establish a reproducible, scalable, and high-purity foundation for embodied manipulation data, providing a robust basis for training high-performance Vision-Language-Action (VLA) models.  

\begin{table*}[t]
\caption{\textbf{Analysis of Open-Source VLA Datasets.} We compare scale, embodiment coverage, strengths, and limitations for commonly used dataset. Here Emb. Adopted only reflects the number of embodiments retained for pretraining after rigorous curation.}
\centering
\small
\begin{tabular}{
  ccccc
  }
\toprule
\textbf{Dataset} & 
\textbf{Scale} & 
\textbf{Emb. Adopted} & 
\textbf{Strengths} & 
\textbf{Limitations} \\
\midrule

OXE~\cite{o2024oxe} & 
$>$1.0M & 
\makecell{11\\Single} & 
\makecell{Large Scale\\Diverse scene} & 
\makecell{Low Complexity\\Short Trajectory\\Poor Quality} \\
% Take for data diversity with strict trajectory filtering. \\
% \makecell{Data Diversity\\Data Cleaning} \\
\addlinespace
\midrule
OXE-AugE~\cite{ji2025oxeauge} & 
4.4M & 
\makecell{9\\Single} & 
\makecell{Inherited from OXE\\Balanced Embodiment} & 
\makecell{Synthetic video\\Visual Artifacts\\Physical Unreality} \\
% Take for embodiment augmentation without fine-grained action modeling. \\
% \makecell{Data Diversity\\Data Cleaning} \\
\addlinespace
\midrule

AgiBot-Beta~\cite{agibotbeta} & 
$>$1.0M & 
\makecell{1\\Dual} & 
\makecell{Long-Horizon\\ Rich Atomic Skill\\ High Quality}& 
\makecell{Single Embodiment} \\
% Control sampling ratio to prevent embodiment bias. \\

\addlinespace
\midrule

RoboCOIN~\cite{wu2025robocoin} & 
180K & 
\makecell{8 \\Single+Dual} & 
\makecell{Hierarchical label\\Diverse Dual-Arm} & 
\makecell{Limited Scale\\No Single-Arm} \\
% Take to enhance complex-task decomposition. \\

\addlinespace
\midrule

RoboMind~\cite{wu2024robomind} & 
107K & 
\makecell{7 \\Single+Dual} & 
\makecell{Unified Protocol\\ Long-Horizon Task\\ Single \& Dual-Arm }& 
\makecell{Nonstandard \\Data Format} \\
% Convert data format to improve cross-embodiment generalization. \\

\addlinespace
\midrule

Galaxea~\cite{jiang2025galaxea} & 
100K & 
\makecell{1\\Dual} & 
\makecell{Long-Horizon\\Subtask Annotation\\Rich Sensor Data} & 
\makecell{Single Embodiment} \\
% Take for fine-grained action representation. \\

\addlinespace
\midrule

\makecell{Others\\(LET~\cite{LET2025},\\Robo360~\cite{liang2023robo360},\\MolmoAct~\cite{lee2025molmoact})} & 
$\le$ 100K & 
-- & 
\makecell{Task-Specific Design}& 
\makecell{Fixed Embodiment\\Inconsist Format\\Adaptation Cost} \\
% Excluded from current pretraining pipeline. \\

\bottomrule
\end{tabular}
\label{tab:dataset}
\end{table*}  

\subsection{Analysis of Open-Source Datasets}  
\label{subsec:data_analysis}

With the rapid advancement of embodied intelligence, an increasing number of high-quality open-source datasets have become available. We systematically integrated the most representative Vision-Language-Action (VLA) data resources currently accessible and conducted a comprehensive evaluation of their characteristics to guide the curation and composition of large-scale pretraining datasets. Details are shown in \Cref{tab:dataset}.
% 随着具身智能的快速发展，越来越多高质量的开源数据集相继出现。我们系统整合了当前最具代表性的视觉-语言-动作（VLA）数据资源，并对其特性进行了全面评估（详细信息见表\Cref{tab:dataset}），以指导大规模预训练数据集的构建与组合。

Existing open-source Vision-Language-Action (VLA) datasets exhibit three key characteristics.
\textbf{Large Scale:} Represented by the Open-X Embodiment (OXE)~\cite{o2024oxe} and OXE-AugE~\cite{ji2025oxeauge} dataset, emphasizing massive data volume and broad coverage of scenes and tasks. These datasets serve as foundational pretraining corpora to foster strong generalization across diverse scenarios and tasks.
\textbf{High Quality:} Exemplified by Agibot~\cite{agibotbeta} and Galaxea~\cite{jiang2025galaxea}, these datasets focus on structured task design, fine-grained annotations, and high physical fidelity. They are well-suited for capability refinement and downstream task adaptation.
\textbf{Embodiment Diversity:} Represented by RoboCoin~\cite{wu2025robocoin} and RoboMind~\cite{wu2024robomind}, these datasets prioritize diversity in robot embodiments (e.g., different morphologies and kinematic structures), constructing data for multiple embodiment types on unified platforms with careful alignment to enhance cross-embodiment generalization.

A single dataset might possess one or two of the above characteristics. For instance, OXE~\cite{o2024oxe} considers both large scale and diversity, Agibot-Beta~\cite{agibotbeta} considers both high quality and large scale, and RoboCoin~\cite{wu2025robocoin} involves both diversity and high quality. However, it is difficult for one dataset to simultaneously satisfy all three characteristics, and there is still much room for improvement.

% 现有的开源VLA数据集涉及到三个方面的关键点/特点：
% 大规模：以Open-X Embodiment（OXE）基准为代表，强调大规模数据量和广泛的场景、任务覆盖，作为基础预训练语料库，促进强大的场景和任务泛化能力。
% 高质量：以Agibot、Galaxea，注重结构化任务设计、细粒度标注和高物理保真度，适用于能力精炼和下游任务适配。
% 本体多样性：以RoboCoin、RoboMind 为代表，注重本体类型的多样性，在统一平台上构建多种类型本体的数据，并进行数据对齐，提升模型的本体泛化性。
% 同一个数据集可能会包含以上1-2个特点，比如OXE同时考虑了大规模和多样性，Agibot同时考虑了高质量和大规模，而RoboCoin同时涉及多样性和高质量，但很难有一个数据同时满足这三个特点，且还有很多上升空间。

To build the dataset for a general-purpose VLA (Vision-Language-Action) model, scale sets the floor, quality defines the ceiling, and diversity determines the scope. Data scale establishes the foundational lower bound of a model's generalization capability. Data quality defines the upper limit of achievable performance. And embodiment diversity determines the boundaries of cross-embodiment generalization. Furthermore, standardized data format such as LeRobot~\cite{cadene2024lerobot} and RLDS~\cite{ramos2021rlds} plays a crucial role in lowering integration barriers and fostering collaborative dataset development within the research community.

% 我们认为对于一个通用VLA模型需要的数据集，规模决定下限，质量决定上限，多样性决定广度”。数据规模决定了模型的泛化能力的基础下限，数据质量则定义了性能可达的上限，而本体多样性决定了跨具身形态泛化的边界。此外，LeRobot和RLDS等标准化数据格式在降低集成门槛、推动研究社区内协作式数据集开发方面发挥着关键作用。

Our dataset currently satisfies these three principles simultaneously through the integration of multiple open-source datasets, data preprocessing (\Cref{subsec:data_process}), and data format and action space alignment (\Cref{subsec:data_format}).
In the future, VLA datasets should strive to deliver higher standard of large scale, high quality, and rich diversity. In terms of scale, emphasis should be placed not only on raw data volume but also on effective interaction duration. Regarding embodiment diversity, datasets should encompass more types of embodiment, including single-arm, dual-arm, semi-humanoid, and full humanoid configurations, to foster cross-embodiment alignment and generalization. For data quality and richness, beyond providing multiple action representations (including both joint and end-effector trajectories) and fine-grained semantic annotations (covering task decomposition, key frames, and object relationships), datasets should also incorporate sensor and dynamics information (such as velocities, forces, and torques), high-definition synchronized multi-view video streams, and camera intrinsics and extrinsics to facilitate the computation of auxiliary 3D geometric labels. These metadata elements serve not only as sources of supervision signals but also as foundational pillars for achieving precise spatiotemporal alignment across vision, language, and action modalities.

% 目前我们的数据集通过整合多种数据集 \Cref{} 、动作空间对齐\Cref{} 来同时满足这三个特点，而未来的VLA数据集,应致力于提供更加大规模、高质量、多样化的数据。规模上不仅要看数据量也要看有效时长。本体多样性上，考虑单臂、双臂、半人形、人形等多种构型，促进多种本体的对齐和泛化。数据质量和丰富性上，除了提供多种动作表示（包括joint与EEF轨迹）和细粒度语义标注（涵盖任务分解、关键帧和对象关系），还可以加入传感器与动力学信息（如速度、力和扭矩）、高清同步多视角视频流、，以及与辅助计算2D/3D几何标签的相机内参和外参。这些元数据元素不仅作为监督信号的来源，更是实现视觉、语言与动作模态之间精确时空对齐的基础支柱。
\begin{figure}[!t]  
    \centering  
    \includegraphics[width=1.0\linewidth]{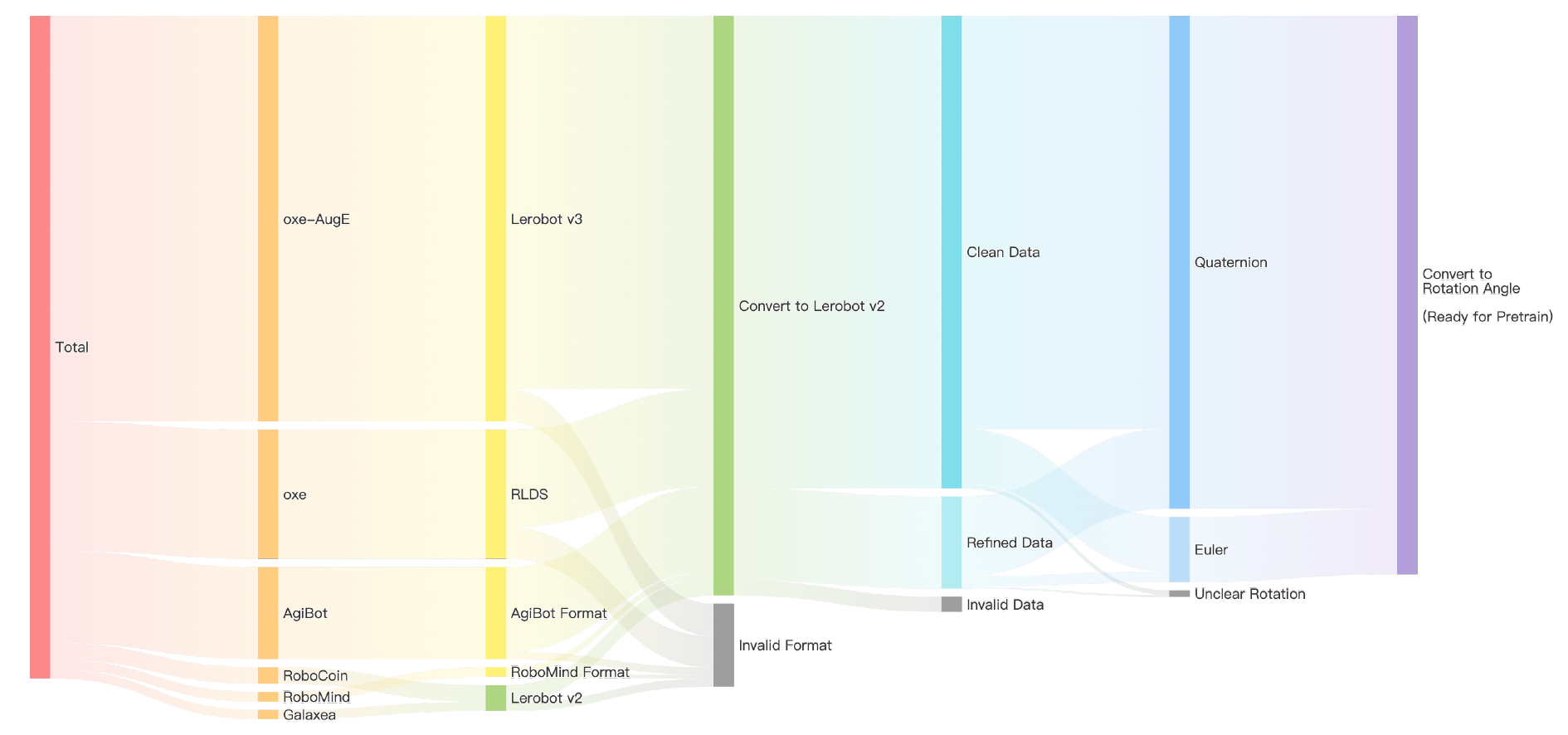}  
    \caption{\textbf{Data cleaning and preprocessing pipeline to construct the UniACT-dataset.}  }  
    \label{fig:data-filter}  
    \vspace{-1em}  
\end{figure}

\subsection{Data Cleaning and Preprocessing}  
\label{subsec:data_process}
%不同数据采用不同策略
According to the analysis in \Cref{subsec:data_analysis}, we take six open-source datasets, including OXE, OXE-AugE, Agibot-Beta, RoboCoin, RoboMind, and Galaxea. As shown in \Cref{tab:dataset}, the overall number of source trajectories is more than seven millions. 
To effectively leverage the unique characteristics and strengths of each dataset, we design dataset-specific curation strategies tailored to their structural and semantic properties. OXE serves as the foundational large-scale single-arm dataset, providing broad task coverage and baseline data diversity. Building upon this foundation, OXE-AugE is incorporated to augment embodiment variation and enrich morphological diversity within the single-arm domain. AgiBot-Beta and Galaxea contribute high-quality visual observations and temporally coherent action sequences. However, given that AgiBot-Beta, despite its substantial scale, features only a single embodiment type, we deliberately reduce its sampling ratio during training to mitigate embodiment bias. Finally, since RoboCOIN and RoboMind are both annotated with fine-grained long-horizon task decompositions across multiple robot morphologies, they are prioritized to strengthen the model's capacity for complex task planning and cross-embodiment generalization for dual-arm robots. This principled integration balances scale, quality, and diversity while actively counteracting distributional imbalances inherent in the raw data sources.

%统一lerobot格式
Despite the abundance of manipulation trajectories provided by multiple open-source datasets, their raw versions often exhibit severe inconsistencies in data format. The raw data originates from heterogeneous sources—including LeRobot v2, LeRobot v3, RLDS (Robotics Learning Data Specification), and various dataset-native formats. To enable unified data processing and loading pipelines, we convert all trajectories into the LeRobot v2 format as a common standard. 

%数据清洗
Moreover, data missing, ambiguous language instructions, and semantic contamination might exist in these datasets. The direct use of such uncurated data risks training models on incorrect language-action mappings or ineffective behavioral policies. To address these challenges, we developed a systematic cleaning and processing pipeline.
Through in-depth analysis of mainstream datasets, we identified several issues. 
% 尽管多个开源数据集提供了大量操作轨迹，但其原始版本普遍存在数据格式不一致、语言指令缺失或模糊、以及语义污染（如占位符文本、截断描述或错位字幕）等问题。直接使用此类未经整理的数据可能导致模型学习到错误的语言-动作映射或无效的行为策略。为此，我们构建了一个系统的清洗流程，对主流数据集的深入分析，发现一些典型问题，比如：
For example, some task prompts contain non-English content (e.g., French, Spanish, and Chinese), nonsensical character sequences, redundant sentences, and even empty content, which severely impair intent parsing and language-action alignment.
For long-horizon tasks, the original task files provide only high-level scene descriptions, while concrete subtask-level instructions are stored in separate files. These require additional processing to decompose trajectories into meaningful subtasks. We emphasize that neglecting proper text alignment will cause the trained model to effectively degenerate into a vision-action (VA) model lacking proper language grounding.
Furthermore, significant frame rate discrepancies exist across OXE subsets—dropping as low as 5 FPS in some cases—leading to ambiguous time steps and increased difficulty in predicting actions for subsequent frames. 
Additionally, certain trajectories exhibit ambiguous or incomplete action annotations. For instance, the semantic meaning of individual action dimensions—such as whether a component encodes translation, rotation (and if so, in which representation), or gripper control—is often unspecified, and some action vectors lack critical components altogether. These deficiencies hinder the model's ability to generate complete and executable action predictions.

% 例如，部分任务提示包含非英语内容（如法语、西班牙语、马达加斯加语、中文）及无意义字符序列，某些轨迹片段甚至存在任务字段为空的情况。Galaxea 数据集中的部分指令还混入了成功评估类语句（例如"任务是否成功？"），严重干扰了意图解析与语言-动作对齐效果。
% 对于长视野任务，原始任务文件仅提供高层场景描述，而具体的子任务级指令则记录在另一独立文件中，需经额外处理方可将轨迹分解为可执行的子任务序列。我们强调，若忽略文本对齐环节，训练所得模型将实质退化为缺乏语言接地能力的纯视觉-动作（VA）模型。
% 此外，OXE 数据集的各子集间存在显著帧率差异（部分子集低至 5 FPS），易导致时序步长混淆，增加下一帧动作预测的难度。

% These issues significantly degrade the effectiveness of language supervision signals and may lead models to incorrectly interpret failed attempts as valid behaviors. 
To mitigate these risks, we designed a multi-stage pipeline to filter out samples with typical issues and refine the remaining data to better align with our requirements. 
% 这些问题显著削弱了语言监督信号的有效性，甚至可能导致模型将失败尝试误判为正确行为。为降低此类风险，我们设计了一个多阶段基于规则的清洗流程，针对以下几类问题样本进行处理：
% 剔除存在典型问题的数据，或者refine数据使其更加适配我们的需求。
Specifically, for \textbf{invalid instructions}, we removed episodes with empty task fields, filtered out garbled instructions with nonsensical sequences, and normalized mixed-language instructions via machine translation to ensure instruction validity and linguistic consistency. 
% 无效指令：移除任务字段为空或含“none”等占位符的片段；过滤包含非英语乱码（如\uXXXX等Unicode转义序列或无意义符号串）的指令；对混合语言指令尝试基于翻译的规范化处理，无法可靠修复的予以舍弃。
To address \textbf{frame-instruction misalignment}, we resolve index mismatches by recomputing temporal alignment across heterogeneous data files. Furthermore, we perform subtask decomposition and insert frame-aligned subtask instructions into the language stream to recover missing granular guidance.
% 数据错位：解决OXE-Aug中episodes.jsonl与tasks.jsonl之间的索引不匹配问题；通过从episodes.jsonl提取action_text并注入帧对齐片段的方式，恢复AgiBot中缺失的子任务指令。
Moreover, we filter out \textbf{visual anomalies}. Specifically, trajectory segments containing visually degraded frames (such as completely black images, severe motion blur, or heavy occlusions) were discarded. Additionally, records captured from ineffective camera viewpoints (e.g. wrist cameras with insufficient field of view to observe the manipulation workspace) were excluded to ensure visual fidelity and task relevance.
% 视觉异常：剔除包含视觉质量差帧（如全黑、严重模糊或高度遮挡图像）的轨迹片段；排除由错位摄像头视角（如腕部摄像头未能覆盖操作区域）采集的记录。
We also check and discard \textbf{abnormal action sequences}. Trajectories exhibiting abnormal length were excluded to avoid failure of original data collection process. And those with large consecutive action deltas were filtered to suppress jitter-induced noise. Furthermore, samples with severe mismatches between action update frequency and video frame rate were eliminated to maintain consistent temporal alignment between modalities.
% 动作异常：过滤连续动作增量异常大（delta > 阈值）的轨迹，以抑制抖动引起的噪声；移除动作更新频率与视频帧率严重不匹配的样本。
For trajectories with incomplete or  \textbf{ambiguous actions}—such as missing action dimensions or unspecified rotation representations (e.g., unclear whether axis-angle, Euler angles, or quaternions are used)—we adopt a strict curation policy and discard such samples to ensure annotation integrity and enable reliable action prediction.
%对于动作规范不完整或存在歧义的轨迹数据——例如动作分量缺失，或旋转表示未明确指定（如无法判断采用轴角、欧拉角还是四元数表示）——我们采取严格的数据筛选策略，直接剔除此类样本，以保障标注完整性并确保模型能够生成可靠的动作预测。

The overall data processing pipeline is illustrated in \Cref{fig:data-filter}. After applying this procedure, approximately 16\% trajectories were discarded. Other trajectories were refined and merged into the final dataset. The remaining high-confidence samples constitute the \textbf{UniACT-dataset}, which serves as the primary source for pretraining our general-purpose VLA model. UniACT-dataset contains more than six million trajectories in 9500+ hours with 20+ embodiments, which is currently the largest collection within the non-private system.
% 整体数据处理流程如图\Cref{fig:data-filter}所示。经过该清洗流程后，约1x%的低质量轨迹被剔除。剩余的六百多万个高置信度样本构成UniACT-dataset数据集，作为我们通用VLA模型预训练的主要数据来源。

\begin{figure}[!t]  
    \centering
    \vspace{-3em}
    \includegraphics[width=1.0\linewidth]{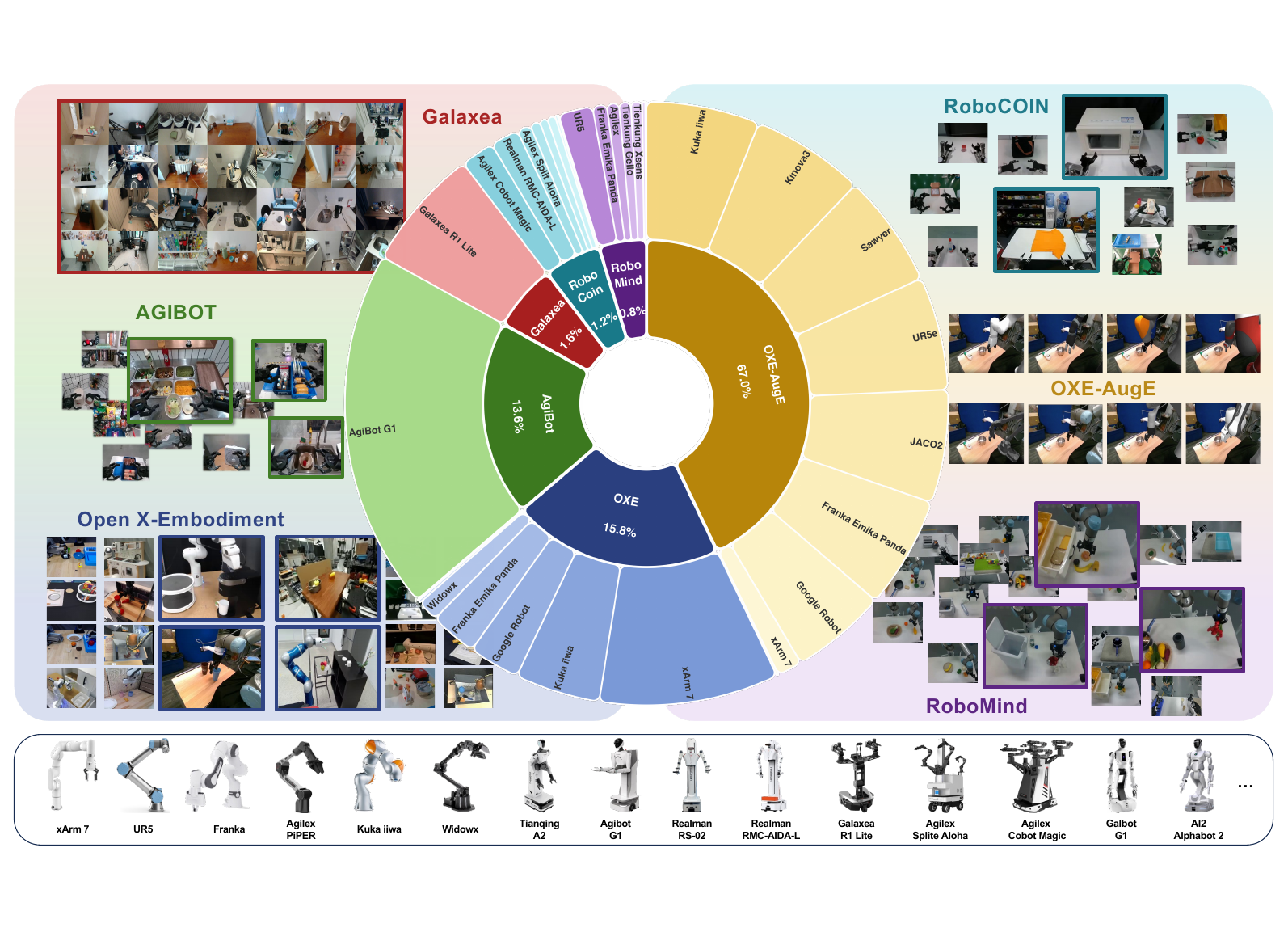}  
    \vspace{-4em}
    \caption{\textbf{Overview of the integrated UniACT-dataset,} which contains more than six million trajectories in 9500+ hours with 20+ unique robot embodiments. }    
    \label{fig:dataset}  
    \vspace{-1em}  
\end{figure}

\subsection{Standardization of Data Formats}  
\label{subsec:data_format}
After \Cref{subsec:data_process}, we have cleaned and preprocessed the trajectories from all sources.
Here in this section we conduct data alignment to ensure full consistency in action space, observation inputs, and training interfaces. This unified framework is built upon three core design principles:  
% 基于LeRobot格式，我们对所有来源的数据进行归一化与重构，确保动作空间、观测输入和训练接口的完全一致性。该统一框架包含三个核心设计原则：

\begin{itemize}  

\item  \textbf{Action Representation: Delta Actions in End-Effector Position with Rotation Vectors.} All tasks are standardized into delta actions with end-effector (EEF) position. Converting absolute actions to relative (delta) actions simplifies and improves the efficiency of model training. And taking EEF rather than joint bridges the embodiment gap to facilitate cross-embodiment generalization. Moreover, all orientation representations (e.g., Euler angles, rotation matrices, quaternions) are transformed into rotation vectors (axis-angle representation), which yield greater stability in action prediction, particularly for fine-grained rotational manipulation.  
The 3D rotation vector of the end-effector is defined as $\bm{r} = \theta \bm{k}$, where $\bm{r} \in \mathbb{R}^3$, $\theta \in [0, \pi]$ is the rotation angle, and $\bm{k} \in \mathbb{R}^3$ (with $\|\bm{k}\| = 1$) is the unit axis about which the rotation occurs.
At each timestep, the policy outputs two 7-dimensional action vectors for both left and right arm, where for each arm the action vector is represented as $[\Delta x, \Delta y, \Delta z, \bm{r}, gripper]$.
This representation enhances controller robustness and ensures cross-platform compatibility. 
% 动作表示：基于末端执行器位置增量与旋转向量的动作编码。 所有任务均被标准化为以末端执行器（EEF）位置为基础的增量动作（delta actions）。将绝对动作转换为相对增量动作可简化训练过程并提升模型训练效率。采用末端执行器空间而非关节空间的动作表示，有助于缩小不同机器人本体之间的形态差异，促进跨本体泛化能力。此外，所有姿态表示（如欧拉角、旋转矩阵、四元数）均被统一转换为旋转向量（轴角表示），该表示在动作预测中展现出更高的稳定性，尤其适用于精细的旋转操作任务。
% 末端执行器的 3D 旋转向量定义为 r=θk，其中 r∈R 3为旋转向量，θ∈[0,π] 为旋转角度（弧度）， （满足 ∥k∥=1）为单位旋转轴。在每个时间步，策略网络为左右机械臂分别输出一个 7 维动作向量，单臂动作向量表示为[Δx,Δy,Δz,r,gripper]，依次对应三维平移增量、3 维旋转向量以及夹爪开合指令。该动作表示方案增强了控制器的鲁棒性，并确保了跨平台的兼容性与泛化能力。

\item \textbf{Unified Single- and Dual-Arm Training via Zero Padding.}  
To allow a single policy network to handle both single-arm and dual-arm tasks, we adopt a pad-to-dual-arm strategy: for single-arm trajectories, the action dimensions corresponding to the unused arm are filled with zeros, and all single-arm data are uniformly treated as right-arm executions within a dual-arm configuration. The model consistently generates dual-arm action sequences but activates only the relevant arm channels during execution. This design enables parameter sharing across tasks while allowing the policy to implicitly learn when to employ one arm versus coordinated bimanual interaction. Guided by task instructions, the model can autonomously determine the appropriate interaction mode during inference, thereby achieving true unification of single- and dual-arm control.  
% 通过零填充实现单臂与双臂训练的统一：为使单一策略网络能够处理单臂和双臂任务，我们采用“补零至双臂”策略：对于单臂轨迹，未使用手臂对应的动作维度用零填充，所有单臂数据统一视为双臂构型中的右臂执行。模型始终输出双臂动作序列，但在执行时仅激活相关手臂通道。该设计促进了任务间的参数共享，同时允许策略隐式学习何时使用单臂或协调双臂交互。在任务指令引导下，模型可在推理过程中自主确定合适的交互模式，实现单臂与双臂控制的真正统一。

\item \textbf{Data Distribution: Diverse Embodiments and Task Scenarios.} The composition of the integrated dataset, including scale proportions and embodiment distribution, is illustrated in \Cref{fig:dataset}.  
The scale and predominant robot embodiments for each dataset are visualized as a pie chart in the center panel. The left and right side illustrate the heterogeneity across datasets in terms of robots, scenes, tasks, camera viewpoints, and visual styles. The integrated dataset with more than six million trajectories spans more than twenty unique robot embodiments and 9500+ hours. At the bottom, we showcase representative robot embodiments including both single-arm and dual-arm. In terms of data distribution, the single-arm dataset OXE-AugE dominates with 67\% of the total volume, followed by OXE as the second largest contributor. The rest four dual-arm datasets collectively account for approximately 17.2\%. Given the inherent scale imbalance and uneven embodiment distribution in the raw data, we employ multi-granularity uniform sampling during training (discussed in \Cref{sec:pretrain}) to balance embodiment coverage with skill learning efficiency.

\end{itemize}  
% \item \textbf{数据分布：多样化的机器人本体与任务场景。} 最终统一后的数据集包含超过 600 万条轨迹片段，来源于六个开源数据集，包括 OXE、OXE-AugE、Agibot-Beta、RoboCoin、RoboMind 与 Galaxea，覆盖广泛的机器人本体构型与任务场景。各数据集呈现出差异化特性。在 \Cref{tab:dataset} 中，我们列出了各数据集的轨迹数量及其所采用的机器人本体类型，并针对不同数据集设计了定制化策略，以有效发挥其独特优势。
% 数据集的构成情况（包括规模比例与本体分布）如 \Cref{fig:dataset} 所示。中心饼图展示了各数据集的规模量级及其主导机器人本体类型；左右两侧呈现了各数据集在机器人平台、场景环境、任务类型、相机视角与视觉风格等方面的多样性。整合后的数据集涵盖二十余种独特的机器人本体，底部展示了代表性本体形态，包括单臂与双臂构型。在数据分布方面，单臂数据集 OXE-AugE 占比最高（67%），其次为 OXE；其余四个双臂数据集合计约占 17.2%。针对原始数据中存在的规模不均衡与本体分布不均问题，我们在训练过程中采用多粒度均匀采样策略（详见 \Cref{sec:pretrain}），以在保障本体覆盖广度的同时提升技能学习效率。

\section{The \VLAFM{} Model}
\label{sec3}

\begin{figure*}[t]
    \centering
    \includegraphics[width=1.0\linewidth]{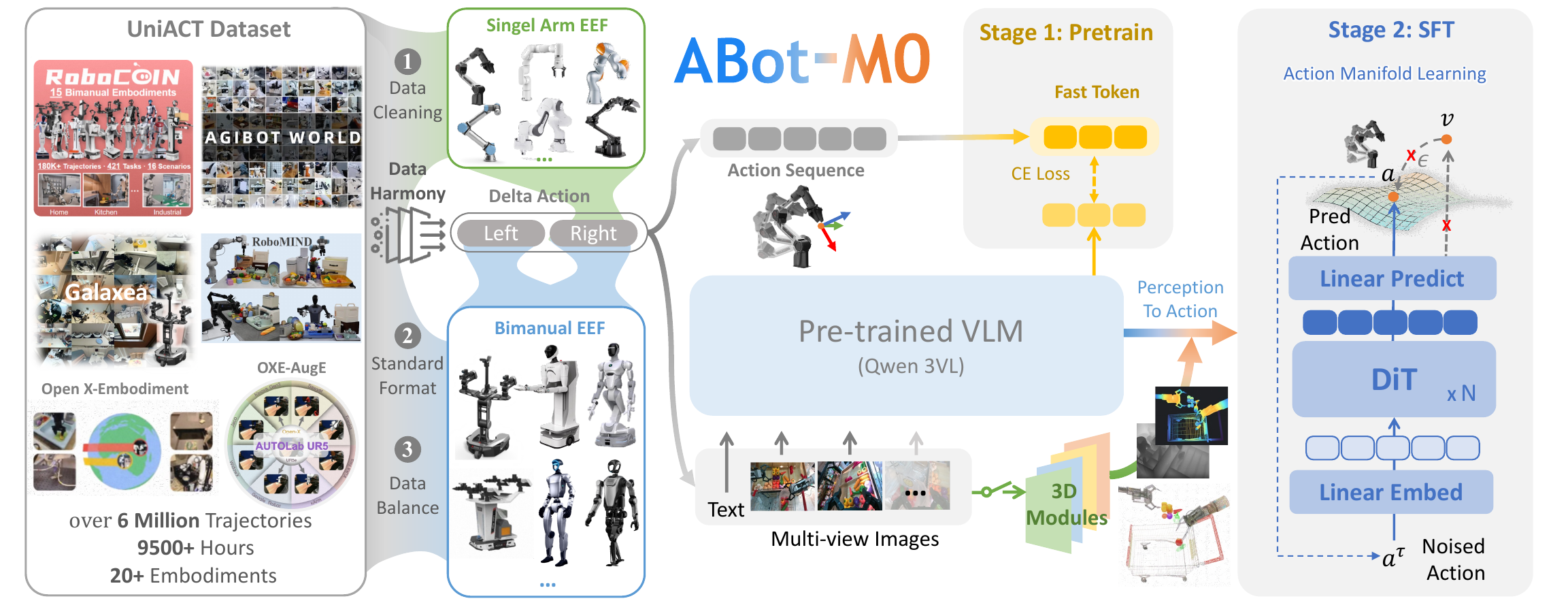}
    \caption{\textbf{Model architecture of \VLAFM{}.} We employ a two-component architecture consisting of a VLM and an action expert. In addition, we utilize action manifold learning to predict actions with two-stage training paradigm. We then carefully select VLM features and further introduce an optional 3D module and  to enhance spatial reasoning.
    }
    \label{fig:model}
\end{figure*}

This section presents the detail of \VLAFM{} model. The overall structure is shown in \Cref{fig:model}. To achieve an end-to-end mapping from multimodal perception to robot action generation, as discussed in \Cref{subsec:Model_Architecture}, we employ a two-component architecture consisting of a VLM and an action expert. In addition to the basic model, we utilize \textit{action manifold learning} to directly predict actions. \Cref{subsec:2stage_training} introduces the two-stage training paradigm. The first stage performs unified pre-training on large-scale heterogeneous data to learn general, cross-task and cross-embodiment manipulation policies. The second stage introduces fine-grained spatial priors through Supervised Fine-Tuning (SFT) to enhance execution accuracy on complex tasks while preserving the model's generalization capabilities.
% 通向通用具身智能的核心挑战在于：如何使模型不仅能理解任务意图，还能有效感知环境结构？
% 视觉语言模型（VLMs）擅长解析指令、识别物体并理解上下文语义。然而，其对三维空间的感知往往停留在语义相对定位层面——例如知道“杯子在盒子左侧”，却无法判断具体偏移距离或是否处于可操作范围内。相比之下，机器人动作执行要求毫米级的空间推理能力与动态协调性。因此，仅依赖 VLM 难以支撑精细操作任务。
% 为此，我们提出一种双流特征协作架构，将 VLM 提供的语义流与多样化基础模型生成的 3D 空间流视为动作生成的两大互补支柱。这两条流并非孤立输入，而是在统一表征空间内进行层次化交互与动态融合，共同引导动作专家（Action Expert）做出有依据且精准的决策。
We further introduce a feature collaboration architecture in ~\Cref{subsec:Percept_2act} with an optional 3D spatial module to enhance spatial reasoning and dynamic coordination.

% 本节介绍我们的模型 \VLAFM{}。为了实现从多模态感知到机器人动作生成的端到端映射，我们采用“VLM + 动作专家”两组件架构，以及一个可选的3D空间模块（见章节6.3.3），最后我们采用动作流形学习以预测干净的动作本身。同时我们构建了两阶段训练范式:第一阶段在大规模异构数据上进行统一预训练，学习跨任务与跨本体的通用操作策略；第二阶段通过监督微调（SFT）引入精细的空间先验，在提升复杂任务执行精度的同时保持模型的泛化能力。

\subsection{Model Architecture}
\label{subsec:Model_Architecture}
\paragraph{Visual Language Model Backbone}
Our model separates vision-language understanding from action generation into two specialized components, including a Visual Language Model (VLM) as the perception engine and an action expert as the decision module. The VLM processes stacked multi-view image sequences, typically captured from front-facing, wrist-mounted, and top-down cameras, together with natural language instructions. Both modalities are independently tokenized and fused into a unified token sequence to enable cross-modal reasoning. We use Qwen3-VL as the backbone VLM due to its strong image-text alignment and long-sequence modeling capability, which is one of the few open-source models that combine high performance with broad applicability. By jointly processing visual and linguistic inputs, the VLM produces spatially aligned multimodal representations, including rich visual features and textual embeddings, which are passed to the action expert as context for action prediction.
% 视觉语言模型骨干. 我们的模型将视觉语言理解与动作生成分离为两个专门但协同工作的模块：作为感知引擎的视觉语言模型（VLM），以及作为决策模块的动作专家。VLM 处理多视角图像序列——如前视、腕部和俯视图——以帧堆叠形式输入，同时结合自然语言指令，这些指令被分词后与视觉标记拼接。我们选用 Qwen3-VL 作为主干 VLM，因其具备强大的图文对齐能力、长序列建模优势及良好的中文支持，是少数兼具高性能与广泛适用性的开源模型之一。通过联合处理视觉与语言输入，VLM 生成空间对齐的多模态表示，包括丰富的视觉特征和文本嵌入，传递给动作专家作为动作预测的上下文。

\begin{figure*}[t]
    \centering
    \includegraphics[width=1.0\linewidth]{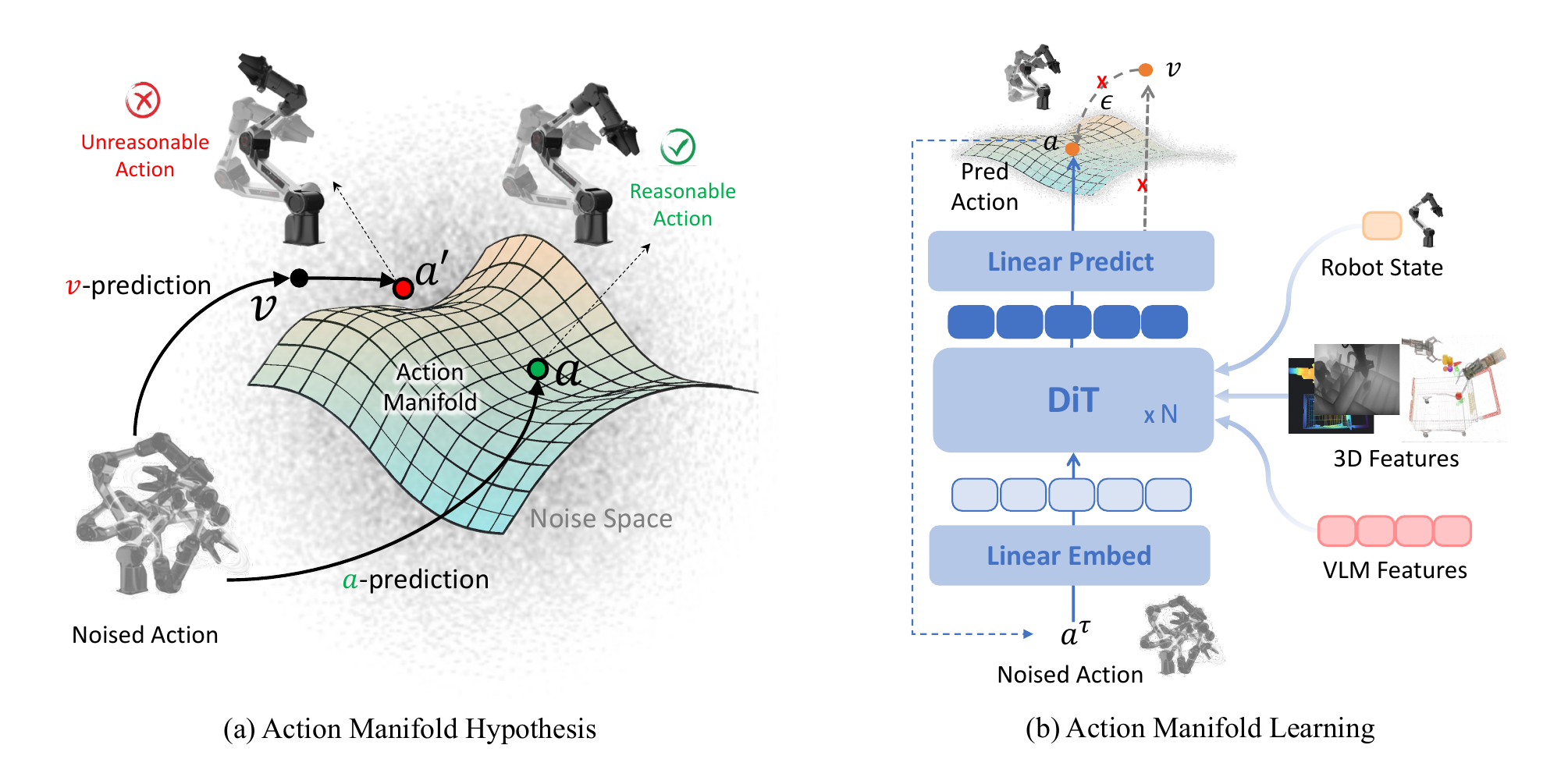}
    \caption{\textbf{Action Manifold}. (a) We posit that a meaningful action sequence is a highly structured entity residing on a low-dimensional action manifold. The conventional prediction targets of noise or velocity are inherently high-dimensional and off-manifold, which increase the burden of model learning and lead to unreasonble action. (b) We propose to predict the action directly rather than velocity, which enables the model to focus on learning the intrinsic structure and semantics of actions.
    }
    \label{fig:jit}
\end{figure*}

\paragraph{Action Manifold Learning}
Conventional diffusion or flow-based generative models are typically trained to predict noise ($\epsilon$-pred) or velocity ($v$-pred), as illustrated in \Cref{fig:jit}. We argue that this poses a fundamental limitation for robot learning. According to the manifold hypothesis~\cite{SSL,carlsson2009topology}, real-world data such as natural images and human language do not scatter randomly across their high-dimensional representation space but rather lie on an intrinsic low-dimensional manifold. We extend this insight to robotics, positing that an effective, coherent, and meaningful sequence of robot actions is also a highly structured entity residing on a low-dimensional \textit{action manifold}. In stark contrast, the prediction targets of noise or velocity are inherently high-dimensional and off-manifold~\cite{JIT,vincent2010stacked}. Forcing a network with finite capacity to directly regress these unstructured, high-dimensional targets is inefficient. The challenge escalates significantly as the embodiment complexity grows from a single robotic arm to a full-body humanoid with two dexterous hands, which results in a substantially expanded action space and increased degrees of freedom. To relieve the pressure of model learning, we propose to predict the action directly rather than velocity, marked as $a$-pred. This shift enables our model, particularly the action expert, to focus on learning the intrinsic structure and semantics of actions, rather than expending its capacity on the inefficient task of filtering out noise from the vast ambient space.
% 在构建我们的模型架构时，我们进行了一项关键的、源于第一性原理的范式迁移。传统的基于扩散或者流的生成模型通常将网络训练目标设定为预测噪声或流速度。然而，我们认为，对于机器人学习而言，这是一个根本性的局限。根据经典的流形假设，真实世界的数据（如自然图像、人类语言）并非随机散布于其高维表征空间中，而是优雅地存在于一个内在的低维流形上。我们将这一洞察延伸至机器人领域：一个有效、连贯且有意义的机器人动作序列，本质上也是一个高度结构化的实体，它同样遵循着一个低维的“动作流形”。与此相对，噪声(ϵ-pred)或混合了噪声的流速度(v-pred)，其本质是高维且遍布整个空间的（off-manifold）。强迫一个能力有限的神经网络去直接拟合这些高维、无结构的目标，不仅低效，而且在动作维度急剧增加时（例如，从七自由度臂到灵巧手乃至全身人形机器人），模型容量会成为巨大瓶颈，甚至可能导致灾难性的失败。为此，我们回归本源，让模型直接预测其最擅长、最本质的目标——纯净的、位于流形上的动作本身。我们将此范式称为动作预测（a-pred）。这一转变使得我们的模型，尤其是其核心的Transformer模块，能够专注于学习动作的内在结构与语义，而非在庞大的高维空间中“大海捞针”般地滤除噪声。

We employ the Diffusion Transformer\cite{Peebles2022DiT} (DiT) as the action generator, marked as $V_\theta$. We apply flowing matching for action learning. Instead of predicting flow velocity, the action generator predicts the denoised action chunk $\hat{{A}}_t = [\hat{{a}}_t, \hat{{a}}_{{t+1}}, \dots, \hat{{a}}_{{t+H-1}}]$ in length of $H$.
Given a ground-truth action chunk $A_t$, a diffusion timestep $\tau \in [0, 1]$, and noise $\epsilon$ sampled from a standard normal distribution, we first construct the noisy action $A^\tau_t = \tau A_t + (1 - \tau) \epsilon$. The network $V_\theta$ then takes the features $\phi_t$ from previous models including VLM and 3D modules, the current robot state $q_t$, and the noisy action $A^\tau_t$ as input and directly predict the action chunk $\hat{{A}}_t$ as the estimation of ground-truth action $A_t$:

\begin{equation}
    \hat{{A}}_t = V_\theta(\phi_t, A^\tau_t, q_t).
\end{equation}

Although the model directly predicts action chunk $A_t$, here we conduct loss on velocity, which yields better performance than on action both in our experiment  and JiT~\cite{JIT}. 
The estimated velocity $\hat{{v}}$ and ground-truth velocity ${v}$ is derived from

\begin{equation}
\begin{split}
    \hat{v} &= (\hat{{A}}_t - {A}^{\tau}_t)/(1-\tau), \\
    v &= ({A}_t - {A}^{\tau}_t)/(1-\tau)
\end{split}
\end{equation}

Then we calculate the mean squared error (MSE) loss on velocity, which equals to a reweighted form of the action loss:

\begin{equation}
\mathcal{{L}}(\theta) = \mathbb{{E}} \| v_{{\text{{pred}}}} - v_{{\text{{target}}}} \|^2  = \mathbb{{E}}\left[ w(\tau) \| V_\theta(\phi_t, A^\tau_t, q_t) - A_t \|^2 \right],
\label{equal:v}
\end{equation}

where the weight is $w(\tau) = \frac{{1}}{{(1 - \tau)^2}}$. This weighting function stems from the Jacobian of the transformation from the action space to the velocity space. It elegantly preserves the advantage of the flow matching method, which is to dynamically adjust the learning signal's strength across different noise levels. Intuitively, as $\tau$ approaches 1 (low noise), the weight tends to infinity, compelling the model to make fine-grained refinements in the final steps. Conversely, for small $\tau$, the smaller weight allows the model to perform more substantial denoising.

\paragraph{Inference Process} During inference, we follow an Ordinary Differential Equation (ODE) solving trajectory to generate the action. Starting from pure noise $A^0_t \sim \mathcal{{N}}(0, \mathbf{{I}})$, we perform iterative denoising over multiple steps. At each timestep $\tau$, we first use the model to predict the estimated pure action $\hat{{A}}_t = V_\theta(\phi_t, A^\tau_t, q_t)$. And then compute the corresponding instantaneous flow velocity $\hat{{v}}$ according to \Cref{equal:v}.
Finally, we employ a numerical integrator (such as the Euler method or a higher-order solver) to update the action:

\begin{equation}
    A^{{\tau+\Delta\tau}}_t = A^\tau_t + \Delta\tau \cdot \hat{{v}}.
\end{equation}

This update cycle allows us to achieve an elegant action prediction at the model level, while still retaining the smooth and stable trajectory generation capabilities of flow models in the underlying dynamics.

% 我们采用一个基于DiT (Diffusion Transformer) 的网络 $V_{\theta}$ 作为动作生成器，但其预测目标不再是流速度，而是去噪后的纯净动作序列块 $\hat{A}_t = [\hat{a}_t, \hat{a}_{t+1}, \dots, \hat{a}_{t+H-1}]$。
% 给定一个真实的动作块 $A_t$，一个扩散时间步 $\tau \in [0, 1]$ 以及标准正态分布噪声 $\epsilon$，我们首先构造出加噪的动作 $A^{\tau}_t = \tau A_t + (1 - \tau) \epsilon$。然后，$V_{\theta}$ 将视觉-语言特征 $\phi_t$、机器人当前状态 $q_t$ 和加噪动作 $A^{\tau}_t$ 作为输入，直接输出对纯净动作 $A_t$ 的最佳估计：
% $$
% \hat{A}_t = V_{\theta}(\phi_t, A^{\tau}_t, q_t)
% $$
% 尽管模型直接预测 $A_t$，但JiT发现，借鉴流匹配（Flow Matching）的损失函数形式能取得最佳性能。具体而言，我们将原始的流速度匹配损失 $\| v_{\text{pred}} - v_{\text{target}} \|^2$，通过代数转换为一个关于 $A_t$ 的加权均方误差损失：
% $$
% \mathcal{L}(\theta) = \mathbb{E}_{\tau, A_t, \epsilon} \left[ w(\tau) \| V_{\theta}(\phi_t, A^{\tau}_t, q_t) - A_t \|^2 \right]
% $$
% 其中，权重 $w(\tau) = \frac{1}{(1 - \tau)^2}$。这个权重函数源于从动作空间到流速度空间的雅可比行列式，它巧妙地保留了流匹配方法在不同噪声水平下动态调整学习信号强度的优点。直观上，当 $\tau$ 接近1（噪声很小）时，权重趋于无穷大，迫使模型在最后阶段进行精细微调；当 $\tau$ 较小时，权重较小，允许模型进行大幅度的去噪。
% **推理过程**：在推理时，我们依然遵循常微分方程（ODE）的求解路径来生成动作。从一个纯噪声 $A^0_t \sim \mathcal{N}(0, \mathbf{I})$ 开始，我们通过多步迭代去噪。在每一个时间步 $\tau$，我们首先使用模型预测出当前估计的纯净动作 $\hat{A}^{\tau}_t = V_{\theta}(\phi_t, A^{\tau}_t, q_t)$。然后，基于这个预测，我们计算出对应的瞬时流速度 $\hat{v}$：
% $$
% \hat{v} = \frac{\hat{A}^{\tau}_t - A^{\tau}_t}{1 - \tau}
% $$
% 最后，我们使用一个数值积分器（如欧拉法或更高阶的求解器）来更新动作：
% $$
% A^{\tau+\Delta\tau}_t = A^{\tau}_t + \Delta\tau \cdot \hat{v}
% $$
% 通过这个“预测-计算-更新”的循环，我们虽然在模型层面实现了优雅的“动作预测”，但在动力学演化上依然享受了流模型平滑、稳定的轨迹生成能力。

\subsection{Two-Stage Training Paradigm}
\label{subsec:2stage_training}

\paragraph{Stage 1: Large-Scale Pre-training for Generalizable Action Priors}
This stage aims to train the model on massive heterogeneous data so it can capture common manipulation patterns across different tasks and robot embodiments, forming a strong and transferable action prior. We use the \textbf{UniACT-dataset}, which contains approximately 6 million trajectories covering single-arm, dual-arm, and various robotic configurations, ranging from basic grasping to complex interactive behaviors. 
As discussed in \Cref{subsec:data_format}, actions are represented as delta actions in the end-effector (EEF) frame. For each arm, the action vector is represented as $[\Delta x, \Delta y, \Delta z, \bm{r}, gripper] \in \mathbb{R}^7$, where $\bm{r}$ is a three-dimensional rotation vector. For dual-arm format, the action vector is in length of 14. 
The model takes multi-view image sequences and natural language instructions as input and outputs action sequences, aiming to map perceptual signals into executable motor commands. To improve convergence and stability, we apply a discretized modeling strategy using an action classification loss with a fast token head, which quantizes continuous actions in a way that preserves gradient flow during training.
% 该阶段旨在利用海量异构数据训练模型掌握跨任务与跨本体的通用操作模式，建立强健且可迁移的动作先验。我们使用 \textbf{UniACT-dataset} 数据集，包含约 620 万条轨迹，涵盖单臂、双臂及多种机械臂构型，任务范围从基础抓取到复杂交互行为。所有动作均表示为末端执行器（eef）坐标系下的增量动作，即六维相对位姿变化 (Δx,Δy,Δz,Δroll,Δpitch,Δyaw) 加夹爪状态。旋转部分采用旋转向量编码，以避免欧拉角奇点或四元数归一化带来的优化问题。模型接收多视角图像序列与自然语言指令，输出动作序列，目标是将感知信号映射为可执行的控制命令。为加快收敛并提高训练稳定性，我们采用离散化建模策略，结合动作分类损失与快速 token 头，有效量化连续动作，同时保留训练过程中的梯度流动。

To support both single-arm and dual-arm tasks within the same network, we use a \textit{pad-to-dual-arm} strategy: for single-arm tasks, the unused arm's action dimensions are zero-padded, and all such operations are treated uniformly as right-arm actions. The model always generates full dual-arm outputs in length of 14, but only activates the relevant arm-channels during execution. This allows the action expert to share parameters across different embodiments, improving parameter efficiency and enabling the model to infer whether bimanual coordination is needed based on the instruction. Furthermore, to address imbalances in task and embodiment distributions, we apply a dual-weighted sampling strategy based on task category and robot morphology  (detailed in \Cref{sec:pretrain}). Rare tasks are oversampled with probability inversely proportional to their data ratio, while different robot configurations are balanced at the batch level. Experiments show this reduces long-tail bias and improves zero-shot generalization on challenging benchmarks such as RoboCasa and Libero-Plus, providing a solid base for downstream tasks.
% 为在统一网络中支持单臂与双臂任务，我们采用 \textit{pad-to-dual-arm} 策略：在单臂任务中，未使用臂的动作维度补零，所有此类操作统一视为右臂行为。模型始终输出完整的双臂动作序列，但在执行时仅激活对应通道。这使得动作专家能在不同本体间共享参数，显著提升参数利用效率，并使模型能根据指令自主判断是否需要双手协作。为缓解任务与本体分布不均衡问题——例如 pick/place 任务过多而 fold/insert 类技能稀少，或某些平台如 Agibot 占主导地位——我们采用基于任务类别与机器人构型的双重加权采样策略。稀有任务按逆频重采样，不同构型在批次层面均衡采样。实验表明，该策略有效减轻长尾偏差，在 RoboCasa 和 Libero-Plus 等挑战性基准上显著提升零样本泛化性能，为后续精细化调整奠定坚实基础。

\paragraph{Stage 2: Space-Aware Supervised Fine-Tuning via Knowledge Injection}
Although the pre-trained model generalizes well across many tasks, it still exhibits accumulated errors and unstable spatial alignment in high-precision scenarios such as fine insertion, cloth folding, or bimanual cooperation. These tasks require not only semantic reasoning but also accurate modeling of 3D spatial relationships, such as the relative pose between peg and hole, fabric crease directions, or timing between arms. To mitigate this limitation, we conduct supervised fine-tuning (SFT) to incorporate essential 3D spatial priors into the pre-trained model while preserving its generalization capabilities across diverse tasks.
% 尽管预训练模型在多数任务上泛化良好，但在高精度操作场景中仍存在误差累积与空间对齐不稳定的问题，如精细插入、布料折叠或双臂协同作业。这些任务不仅依赖语言理解，还需对三维空间关系进行精确建模，例如销孔之间的相对姿态、布料折痕方向或多臂协调时机。为此，我们进行监督微调（SFT），目标是在不破坏其已有通用行为的前提下，向模型注入关键的三维结构知识。

We fine-tune both the VLM and action expert modules jointly with a small learning rate. Dropout and action noise perturbation are applied during training to increase robustness under real-world conditions. Results show substantial gains in success rates on difficult tasks like insertion, folding, and bimanual door opening, while performance on previously learned simple tasks remains stable. This confirms that new capabilities can be added without degrading existing ones.
% 我们以低学习率（\texttt{xxx}）联合微调 VLM 与动作专家模块，并在训练中引入 dropout 与动作噪声扰动，以增强在真实环境下的鲁棒性。结果表明，在插入、折叠和双臂开门等困难任务上成功率显著提升，而简单任务的表现保持稳定，验证了新能力可叠加而不损害原有能力。

This two-stage approach, comprising large-scale pre-training followed by targeted supervised fine-tuning (SFT), effectively reconciles generality with task-specific precision. The pre-training phase endows the model with a broad understanding of reasonable action spaces, while the subsequent SFT stage specializes the policy for accurate execution under task constraints. 
Critically, this methodology requires no architectural modifications, thereby enabling flexible integration of new skills in a plug-and-play manner. Moreover, the framework supports continual expansion. Emerging sensor modalities such as tactile feedback and force-torque sensing, along with novel task formulations, can be seamlessly incorporated through analogous fine-tuning protocols.

% Overall, the two-stage approach -- pre-training followed by targeted SFT -- achieves a balance between generality and precision. The first stage teaches what counts as a reasonable action; the second refines how to execute it accurately. No changes to the model architecture are required, allowing new skills to be added flexibly. More importantly, this framework supports ongoing development: future sensor modalities (e.g., tactile feedback, force sensing) or new task types can be integrated through similar fine-tuning procedures. Rather than static training, this represents a sustainable path for embodied intelligence to grow in complexity and capability over time.
% 总体而言，这一两阶段方法——预训练加定向微调——在通用性与精确性之间实现了平衡。第一阶段教会模型什么是合理动作，第二阶段则优化如何更准确地执行。整个流程无需修改模型结构，便于灵活扩展新技能。更重要的是，该框架支持持续发展：未来的传感器信号（如触觉反馈、力传感）或新任务类型均可通过类似微调机制集成。这不再是静态的模型训练，而是一条具身智能可持续成长的技术路径。

\section{Pre-Training}  
\label{sec:pretrain}  

A fundamental challenge in advancing general-purpose embodied intelligence centers on distilling invariant representations from large-scale, heterogeneous manipulation datasets that span morphologically diverse robot embodiments and task distributions.
Multi-source embodied datasets exhibit significant variation in trajectory length, task granularity, and embodiment diversity. 
Under naive trajectory-level uniform sampling, the training distribution becomes dominated by data volume, resulting in pathological biases. 
On one hand, long-tail but valuable domains, such as uncommon skills or rare robot morphologies, receive insufficient exposure, hindering effective convergence.  
On the other hand, limited training budgets are consumed by redundant sampling of frequent skills, which limits coverage of rare behaviors and ultimately undermines cross-task and cross-embodiment generalization. 
To address this issue, this section presents a systematic study of sampling strategies for bimanual learning. 
We propose multi-level stratified sampling to balance the embodiment coverage and skill acquisition efficiency, thereby establishing a data-aware trade-off that facilitates robust cross-embodiment and cross-task generalization.
\Cref{subsec:sampling} conducts an in-depth analysis of different sampling strategies and propose a multi-level stratified sampling to balance the embodiment coverage and skill acquisition efficiency.   
\Cref{subsec:eval_sampling} empirically validates how distributional disparities induced by these strategies impact generalization performance. The superiority of Task-Uniform is consistently validated across three dimensions: cross-embodiment generalization, cross-dataset transfer, and downstream task adaptation.

% 实现通用具身智能的核心挑战之一，是从大规模异构操作数据中提取跨形态与跨任务的共性知识。多源具身数据集在轨迹长度、任务粒度和具身多样性方面存在显著差异。在简单的按轨迹均匀采样下，训练分布极易被数据量主导，从而产生病态偏差。一方面，双臂配置或罕见机器人形态等长尾但高价值领域得不到充分暴露，难以有效收敛；另一方面，有限的训练预算被高频技能的重复采样大量消耗，限制了对稀有行为的覆盖，最终削弱跨任务与跨形态的泛化能力。为解决此问题，本节系统研究双臂学习中的采样比例策略。在保留已有单臂训练先验的基础上，引入多粒度均匀采样方法，旨在平衡形态覆盖与技能学习效率，从而识别出支持数据驱动泛化的有效权衡方案。

\begin{figure*}[t]
    \centering
    \includegraphics[width=1.0\linewidth]{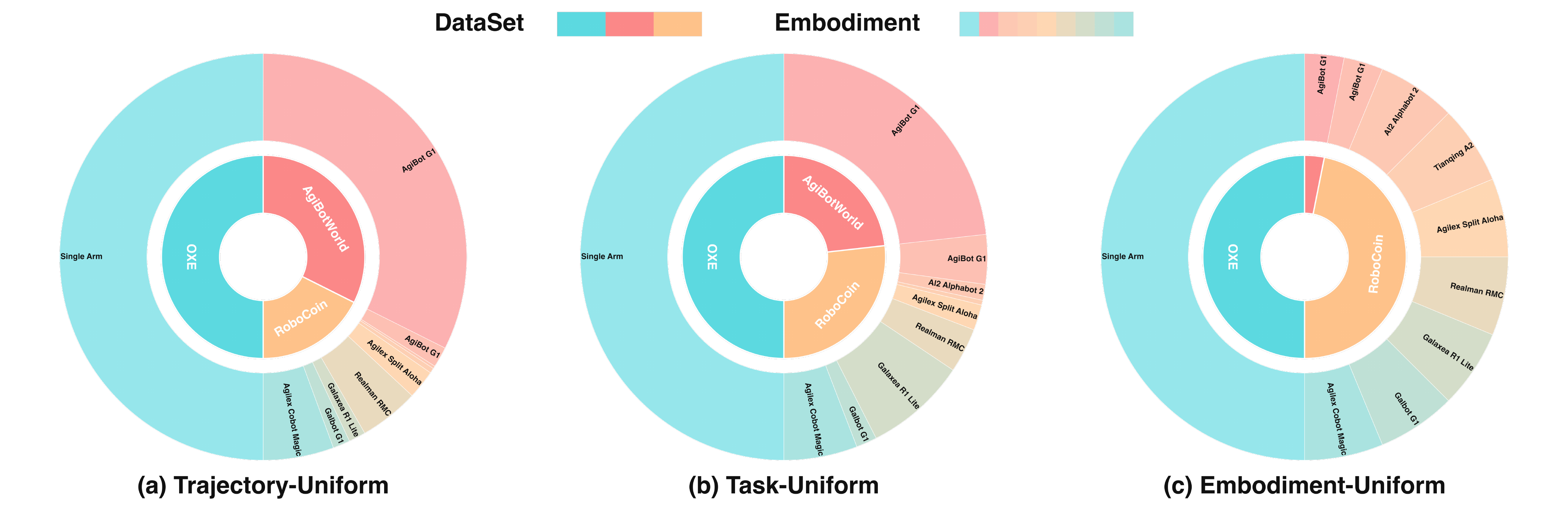}
    \caption{\textbf{Embodiment distribution under different sampling strategies.} Data are drawn from OXE~\cite{o2024oxe}, AgiBot-Beta~\cite{agibotbeta}, and RoboCoin~\cite{wu2025robocoin}. The mixture receipe of single-arm data from OXE follows OpenVLA~\cite{kim2024openvla} and is fixed among these three strategies. We compare (a) Trajectory-Uniform, (b) Task-Uniform, and  (c) Embodiment-Uniform sampling strategies and show the data distribution among different dataset and embodiments.}
    \label{fig:embodiment-distribution}
%\vspace{-1em}
\end{figure*}

\begin{figure*}[t]
    \centering

    % 缩小到 0.64，腾出空间
    \begin{minipage}[t]{0.64\linewidth}
        \centering
        \includegraphics[width=\linewidth]{figure/sampling_ratio.png}
        \caption{\textbf{Skill Sampling Characteristics.} Skill sampling behavior under different bimanual sampling strategies: (a) rank–probability distribution, (b) Lorenz curves of skill sampling probabilities, and (c) Coverage@T, measuring the number of unique skills covered as sampling progresses.}
        \label{fig:sampling-ratio}
    \end{minipage}
    \hfill % 现在 \hfill 会产生约 0.02\linewidth 的间距
    % 缩小到 0.33，确保总和小于 1
    \begin{minipage}[t]{0.33\linewidth}
        \centering
        \includegraphics[width=\linewidth]{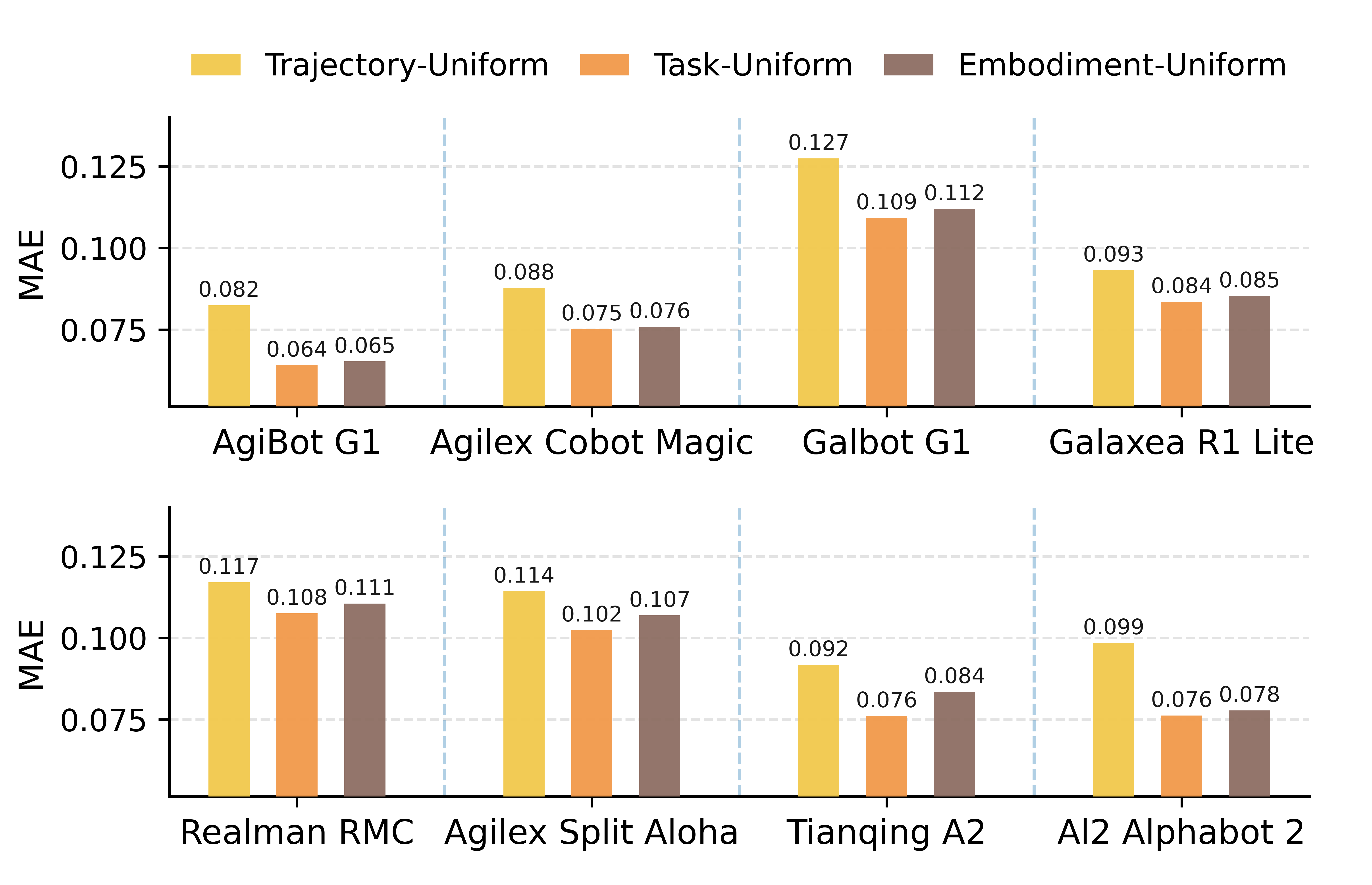}
        \caption{\textbf{Validation Performance Across Bimanual Embodiments on RoboCoin.}}
        \label{fig:embodiment-mae}
    \end{minipage}

    %\vspace{-1em}
\end{figure*}

\subsection{Sampling Ratio for Multi-Embodiment Learning}  
\label{subsec:sampling}
This section systematically analyzes three uniform sampling strategies for bimanual data, focusing on their effects on embodiment distribution and skill sampling behavior in multi-embodiment learning. We fix single-arm data at 50\% of the sampling budget and perform controlled ablation of sampling strategies solely within the bimanual regime.
% 本节系统分析三种双臂数据的均匀抽样策略在构型分布与技能采样行为上的影响，重点关注它们在多构型学习场景下的统计偏置与覆盖效率。单臂数据仅作为背景设置，不再展开讨论。

\textbf{Embodiment-Level Distribution Analysis.} At the embodiment level, the three sampling strategies induce markedly different effective training distributions, as shown in \Cref{fig:embodiment-distribution}. Trajectory-uniform sampling largely preserves the original dataset scale imbalance, causing training to be dominated by AgiBot-G1 and resulting in a highly concentrated embodiment distribution, which increases the risk of homogenization. In contrast, task-uniform sampling substantially alleviates this issue by increasing the sampling visibility of multi-task but single-embodiment data from RoboCoin. This mechanism allows long-tail embodiments to receive more exposure during training, while still retaining AgiBot-Beta as the primary data source, leading to a more stable balance between embodiment coverage and data scale.  
% 在构型层面，不同抽样策略在有效训练数据分布上呈现出显著差异（见 Fig.1）。按轨迹均匀抽样在很大程度上继承了原始数据体量结构，使训练过程明显受 AgiBot-G1 主导，构型分布高度集中，进而增加了同质化风险。相比之下，按任务均匀抽样显著缓解了这一问题：通过提升 RoboCoin 中多任务但单构型数据的采样可见度，长尾构型在训练过程中获得更多曝光，同时仍保留 AgiBotWorld 作为主要数据来源，从而在构型覆盖与数据规模之间取得更为稳定的平衡。

\textbf{Skill-Level Sampling Characteristics.} Apart from embodiment balance, we examine the skill-level sampling behavior of these strategies, shown in  \Cref{fig:sampling-ratio}. Although all strategies exhibit pronounced long-tailed skill distributions, task-uniform sampling produces a noticeably less concentrated probability mass. This is reflected by a Lorenz curve closer to the equality line and a lower corresponding Gini coefficient. Moreover, under the same sampling budget, task-uniform sampling achieves faster growth in the number of covered unique skills, indicating higher efficiency in reducing redundant sampling and improving skill diversity. In contrast, embodiment-uniform sampling, while enforcing stronger balance across embodiments, further concentrates sampling probability on a small set of high-frequency skills, resulting in substantially slower coverage growth.
% 从 skill 维度进一步分析不同策略的采样行为（见 Fig.2），可以观察到三种策略均呈现明显的长尾分布特征。然而，按任务均匀抽样在概率分布层面表现出更低的集中度，其 Lorenz 曲线更接近对角线，对应更小的 Gini 系数。同时，在给定抽样预算下，该策略能够以更快速度覆盖更多不同技能，表明其在减少重复采样、提升技能覆盖效率方面更具优势。相较之下，按构型均匀抽样虽然强化了构型层面的均衡性，但其 skill 采样概率进一步向少数高频技能集中，导致技能覆盖效率显著下降。

Overall, task-uniform sampling provides a more favorable trade-off between embodiment diversity and skill coverage, whereas trajectory-uniform sampling suffers from severe scale dominance and embodiment concentration, and embodiment-uniform sampling, despite improving embodiment balance, introduces stronger bias toward high-frequency skills and degrades skill coverage efficiency.

% 总体而言，按任务均匀抽样在构型多样性与技能覆盖之间提供了更为有利的折中；相比之下，按轨迹均匀抽样受数据规模主导，容易造成构型分布集中，而按构型均匀抽样尽管提升了构型层面的均衡性，却引入了更强的高频技能偏置，从而降低了技能覆盖效率。

\begin{figure}[tbp]
  \centering
  % --- 左侧图片：占用 2/3 宽度 ---
  \begin{minipage}[b]{0.66\textwidth}
    \centering
    \includegraphics[width=\linewidth]{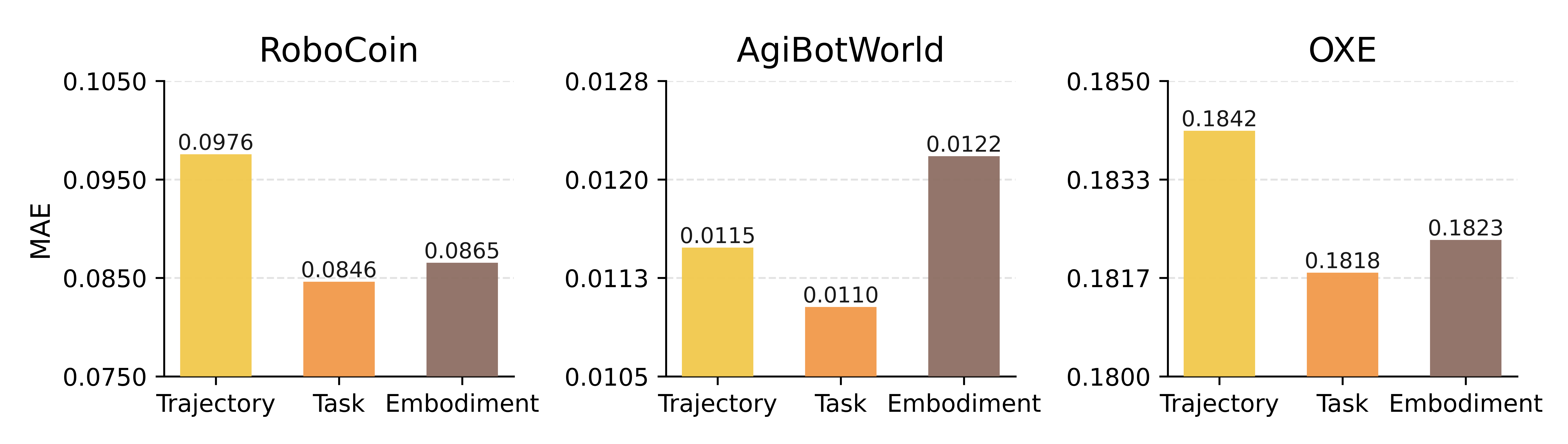}
    \caption{\textbf{Dataset-Wise Validation Performance Under Different Sampling Strategies.}}
    \label{fig:dataset-mae}
  \end{minipage}
  \hfill
  % --- 右侧表格：占用 1/3 宽度 ---
  \begin{minipage}[b]{0.32\textwidth}
    \centering
    % 先放表格内容
    \resizebox{\linewidth}{!}{%
      \begin{tabular}{lc}
      \toprule
      \textbf{Method} & \textbf{Total} \\
      \midrule
      Trajectory-Uniform & 71.3 \\
      Embodiment-Uniform & 71.6 \\
      \textbf{Task-Uniform}       & \textbf{72.4} \\
      \bottomrule
      \end{tabular}
    }
    \vspace{13pt}
    % 将 Caption 移至下方
    \captionof{table}{\textbf{Downstream Performance on Libero Plus.}}
    \label{tab:downstream-liberoPlus}
  \end{minipage}
\end{figure}

\subsection{Generalization Evaluation} 
\label{subsec:eval_sampling}

In \Cref{subsec:sampling}, we have analyzed how different bimanual sampling strategies introduce structural biases during training from the perspectives of embodiment distribution and skill sampling statistics. To further examine whether these differences translate into measurable generalization behavior and downstream performance, this section conducts a systematic evaluation across datasets and embodiments. Notably, pretraining is conducted exclusively on OXE, AgiBot-Beta, and RoboCoin, while Libero is used only for downstream supervised fine-tuning.

% 在前一节中，我们从构型分布与技能采样统计特性出发，分析了不同双臂数据配比策略在训练阶段引入的结构性偏置。为进一步验证这些差异是否会实质性影响模型的泛化能力与下游表现，本节通过跨数据集与跨构型的定量评测，对三种配比策略进行系统比较。需要指出的是，预训练阶段仅使用 OXE、AgiBotWorld 与 RoboCoin 三个数据集；Libero 仅用于下游监督微调与评估。

\textbf{Validation Set Construction.} To avoid validation results being dominated by dataset scale or partition granularity, the validation set is constructed by aligning the sampling unit with the most representative diversity dimension of each dataset. Specifically, for OXE, since the single-arm sampling ratio is fixed, we randomly sample 1,000 trajectories as the validation set. For AgiBot-Beta, we stratify by task and randomly sample one trajectory for each task (183 in total) to ensure task coverage. For RoboCoin, we stratify by embodiment and randomly sample 30 trajectories per embodiment (240 in total) to ensure embodiment coverage. The evaluation metric is the mean absolute error (MAE) between the predicted and ground-truth actions. All models are trained under identical settings for 50k steps before evaluation. 
% 为避免验证结果被单一数据集的体量规模或划分粒度主导，验证集的构建遵循“抽样单位对齐各数据集最具代表性的多样性维度”的原则。具体而言，对 OXE 随机抽取 1000 条轨迹（trajectory），由于其训练配比固定，验证集不刻意偏向任务或构型；对 AgiBotWorld 按任务（task）分层，每个任务随机抽取 1 条轨迹（共 183 条），以保证任务覆盖；对 RoboCoin 按构型（embodiment）分层，每个构型随机抽取 30 条轨迹（共 240 条），以保证构型覆盖。评价指标采用预测动作与真实动作之间的平均绝对误差（MAE）。在完全相同的训练设置下，三种配比策略均训练至 50k steps 后进行评估。

\textbf{Cross-Embodiment Generalization.} We assess cross-embodiment generalization using the RoboCoin validation set stratified by bimanual embodiment. Results are shown in \Cref{fig:embodiment-mae}. Task-Uniform and Embodiment-Uniform achieve similar overall MAE, while Task-Uniform is consistently lower on most embodiments. In contrast, Trajectory-Uniform performs substantially worse, exhibiting markedly higher MAE across nearly all embodiments. These results suggest that task-level organization provides more transferable interaction supervision, enabling strong cross-embodiment performance without enforcing strict embodiment-level uniformity.
% 我们以 RoboCoin 验证集中不同双臂构型的评测来刻画跨构型泛化能力（见 Fig. 2）。Task-Uniform 与 Embodiment-Uniform 的整体 MAE 接近，但 Task-Uniform 在多数构型上进一步取得更低误差。相比之下，Trajectory-Uniform 在几乎所有构型上显著更高，表现出明显的跨构型泛化短板。这一结果表明，按任务粒度组织数据能够带来更可迁移的交互监督信号，即使不对构型分布施加严格均匀约束，也能实现与 Embodiment-Uniform 相当甚至更优的跨构型性能。

\textbf{Cross-Dataset Generalization.} Results in \Cref{fig:dataset-mae} indicate that Task-Uniform achieves the lowest MAE on OXE, AgiBot-Beta, and RoboCoin, demonstrating a more stable and consistent advantage. In contrast, Trajectory-Uniform yields the highest errors on RoboCoin and OXE, while Embodiment-Uniform performs worst on AgiBot-Beta, reflecting the biases induced by different uniformity objectives under cross-dataset evaluation. This suggests that Task-Uniform better balances learning signal and coverage needs across data sources, leading to a more favorable overall trade-off in cross-dataset generalization.
% 在三数据集的整体评测中（见 Fig. 1），Task-Uniform 在 OXE、AgiBotWorld 与 RoboCoin 上均取得最低 MAE，体现出更稳定的一致性优势。相比之下，Trajectory-Uniform 在 RoboCoin 与 OXE 上误差最高，而 Embodiment-Uniform 在 AgiBotWorld 上误差最高，反映出不同均匀化目标在跨数据集场景下带来的偏置差异。这说明 Task-Uniform 能更好地在不同数据源之间权衡学习信号与覆盖需求，从而在跨数据集评测中表现出更优的整体折中。

\textbf{Downstream Transfer on Libero Plus.} Beyond MAE evaluation, we further transfer the pretrained models to the Libero Plus benchmark for supervised fine-tuning to assess downstream transferability. As reported in \Cref{tab:downstream-liberoPlus}, we observe noticeable performance differences across sampling strategies, and the overall trend is consistent with the MAE evaluation. These results suggest that the choice of sampling strategy during pretraining can influence downstream learning behavior and final performance on manipulation tasks.
% 除 MAE 评测外，我们进一步在 Libero 上进行监督微调（SFT），并在主流下游基准 Libero Plus 上评测，以检验不同配比策略在实际下游任务中的迁移效果。如表 1 所示，不同配比策略在下游任务上同样呈现稳定差异，且与前述 MAE 评测的整体趋势保持一致，表明预训练阶段引入的配比差异会直接影响下游操作任务的学习质量。

\section{Perception to Action}
\label{subsec:Percept_2act}

% A fundamental challenge in advancing general-purpose embodied intelligence lies in endowing models with dual capabilities: comprehending high-level task intent while simultaneously grounding actions in precise environmental structure.
A fundamental challenge in advancing general-purpose embodied intelligence lies in endowing models with dual capabilities: comprehending high-level task intent while simultaneously grounding actions in precise environmental structure.
%
% Vision-Language Models (VLMs) excel at parsing instructions, recognizing objects, and understanding contextual semantics. However, their perception of 3D space often remains at the level of semantic relative positioning—knowing that “the cup is to the left of the box,” but not how far left or whether it is within reach. 
Vision–language models (VLMs) demonstrate strong capabilities in parsing natural language instructions, recognizing object categories, and inferring contextual semantics. Nevertheless, their spatial perception typically remains qualitative, capturing only semantic relations such as "the cup is to the left of the box" without metric precision regarding distance, reachability, or egocentric pose. 
In contrast, robotic action execution demands millimeter-level spatial reasoning and dynamic coordination. Therefore, relying solely on VLMs is insufficient for fine-grained manipulation.
To address this, we discuss the choice of VLM feature for action expert in \Cref{subsec:vlm_feat} and compare different mechanisms to inject 3D information in \Cref{subsec:3dinfo}.
VLM-derived visual features and geometry-aware 3D representations undergo hierarchical cross-attention within a shared latent space, producing a fused multimodal embedding that conditions the action expert to generate spatially grounded and temporally coherent action sequences.

% 通向通用具身智能的核心挑战在于：如何使模型不仅能理解任务意图，还能有效感知环境结构？
% 视觉语言模型（VLMs）擅长解析指令、识别物体并理解上下文语义。然而，其对三维空间的感知往往停留在语义相对定位层面——例如知道“杯子在盒子左侧”，却无法判断具体偏移距离或是否处于可操作范围内。相比之下，机器人动作执行要求毫米级的空间推理能力与动态协调性。因此，仅依赖 VLM 难以支撑精细操作任务。
% 为此，我们提出一种双流特征协作架构，将 VLM 提供的语义流与多样化基础模型生成的 3D 空间流视为动作生成的两大互补支柱。这两条流并非孤立输入，而是在统一表征空间内进行层次化交互与动态融合，共同引导动作专家（Action Expert）做出有依据且精准的决策。

\subsection{VLM Feature Interaction}
\label{subsec:vlm_feat}
\begin{figure*}[t]
    \centering
    \includegraphics[width=1.0\linewidth]{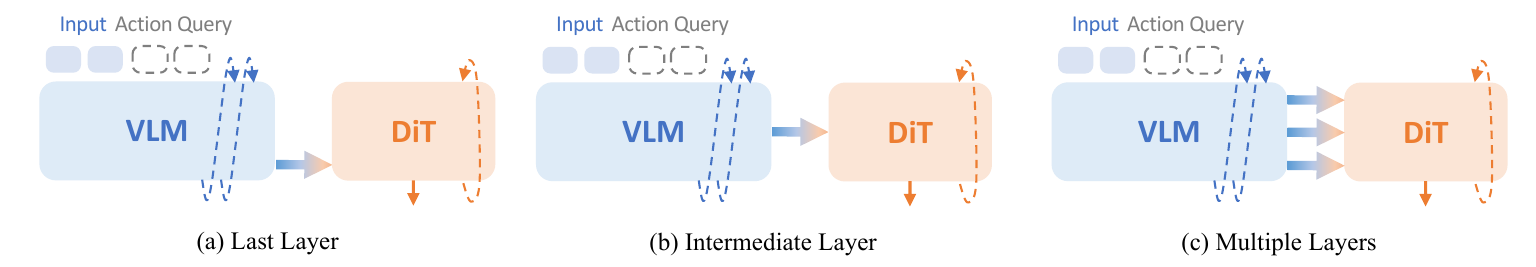}
    \caption{\textbf{VLM Feature Interaction.} Following pre-training on robotics data, we feed an action expert with either the VLM's raw hidden features or an additional action query, extracted from its final, intermediate, or multiple layers.    }
    \label{fig:vlm_feature}
%\vspace{-1em}
\end{figure*}

Our vision-language model (VLM) is pre-trained on large-scale vision-language-action (VLA) data, endowing it with strong generalization capabilities in embodied scenarios and the ability to directly predict discrete actions in textual form. However, architectural, training, and output-format differences persist between the pre-trained VLM and the downstream action expert policy model, such as the discrete versus continuous action representation. In contrast to prior work like VLA-Adapter~\cite{wang2025vlaadapter} that focused on feature selection for VLMs without robotics pretraining, we explore which features from a VLM pretrained with robotic data are most critical for training an action expert.
%我们的VLM模型会经过大规模VLA数据的预训练，使得VLM具备了具身场景的泛化能力，且能直接通过文本形式预测离散action。但VLM和action expert的policy 模型设计仍然存在差异，无论是模型结构、训练方式还是输出格式（离散vs连续）都不同。与之前专注于为无机器人预训练的 VLM 进行特征选择的研究（如 VLA-Adapter）不同，我们探讨了在经过机器人预训练的 VLM 中，哪些特征对于训练动作专家最为关键。

We conduct ablation studies with Qwen3-VL-4B as the VLM backbone and a 16-layer Diffusion Transformer (DiT) as the action expert. We explore two key aspects of feature utilization: (a) Which layer(s) of the pre-trained VLM provide the most effective features for policy learning?
We evaluate features from three different layers, including features of final-layer(deepest representation), features of intermediate-layer (approximately the 70th percentile layer in depth), and concatenated features from all hidden layers. 
(b) Does incorporating action queries yield better performance than using raw VLM features alone? We also experimented with action queries drawn from the intermediate layer, the final layer, and the last 16 layers, respectively. Furthermore, we investigated using the concatenated features of the query and the hidden states from the last 16 layers of VLM as a condition.
The schematic illustration is provided in \Cref{fig:vlm_feature}.

%在实验中，我们采用Qwen3VL-4B和16层的DiT做对比实验。我们主要关注两个问题：
% 问题 1：预训练后，VLM 中的哪一层或者哪几层特征对策略网络更有效？高层特征还是低层特征更有帮助？
% 我们做了三种实验：
% 1. VLM的最后一层特征
% 2. VLM取中间层，取中间70%的一层隐藏特征
% 3. VLM所有隐藏层
% 问题 2:ActionQuery 功能是否比原始的VLM特征功能更好？
% 我们测试了两种方案：
% 我们也分别对中间层，最后一层和后16层的action query进行了实验。最后，我们也探索了将后16层的隐藏层特征和query拼接作为条件。

Experimental results shown in \Cref{tab:vlm_libero_plus} (detailed in ablation study) lead to the following conclusions:
(i) Using raw VLM features directly outperforms feature concatenation with action queries.
(ii) Deep-layer VLM features consistently yield better policy performance than shallow or intermediate features.
(iii) Aggregating features from multiple layers provides no significant improvement over using the final layer alone.

%实验结果见下表，具体细节见消融。
% 实验结论：
% 1、直接采用VLM特征，比拼接action query效果更好。
% 2、VLM的深层特征比浅层特征更有效。
% 3、多加入VLM特征层数，并不会带来更多收益。

These findings suggest that large-scale VLA pre-training enables the VLM to internalize action-space semantics, where intermediate layers capture rich multi-modal scene representations and deeper layers progressively encode action-relevant semantic structures. Consequently, the model no longer requires auxiliary mechanisms such as action queries to adapt the VLM to action prediction tasks. The superior performance of the final-layer features surpasses  both shallow-layer and multi-layer alternatives and demonstrates that a single deep representation effectively encapsulates the most discriminative action-space information. Moreover, the under-performance of action-query augmentation further confirms that the pre-trained VLM has already achieved sufficient alignment with the action domain, rendering additional adaptation modules redundant.

This study not only validates the effectiveness of our VLA pre-training paradigm but also provides a practical and efficient feature selection strategy for transferring pre-trained VLMs to downstream action generation modules—favoring simplicity (single deep-layer features) over complexity without sacrificing performance.
%我们的model经过大规模VLA数据的预训练，具备了具身场景的泛化能力，且能直接通过文本形式预测离散action。这意味着预训练后的VLM已经能够提取一定动作空间特征，中间层能够提取具身场景的多模态特征，越深层的特征越接近预测的动作空间语义特征，也不再需要额外的action query去帮助VLM适应动作空间。实验结果也证明了，直接用预训练后VLM的最后一层深层特征，比使用浅层特征效果更好，比使用所有层的效果也更好，相当于一层最深层特征，学到了最有效的动作空间特征。并且，action query在预训练后的VLM中也并没有发挥效果，甚至更差，因为模型已经不需要额外的query来辅助VLM快速迁移到VLA任务。这部分的实验间接证明了，我们预训练的有效性。也为预训练后VLM的特征选择提供了一种有效的方案。

\subsection{3D Information Injection}
\label{subsec:3dinfo}
While \Cref{subsec:vlm_feat} establishes that deep-layer VLM features alone already encode rich action-relevant semantics, they remain inherently limited by their 2D visual input. VLMs excel at semantic understanding but lack precise geometric awareness. It knows ``where'' but can not tell ``how far'' or ``whether reachable''. This gap becomes critical in precise manipulation, where spatial reasoning is required.

To complement the VLM's semantic feature with explicit 3D spatial structure, we introduce a plug-and-play 3D information injection module that operates alongside the VLM within our dual-stream architecture. Rather than replacing the VLM, this module enriches its representation with geometric priors, enabling the unified system to jointly reason about what to do and where to act.

Specifically, we integrate two sources of 3D awareness.
(i) Feedforward single-image 3D features: We use a model pretrained on large-scale 3D data, such as VGGT~\cite{wang2025vggt}, to extract 3D-aware features from a single RGB image. By jointly modeling appearance and geometry, VGGT infers the structure of the 3D environment even from monocular input.
(ii) Implicit multi-view features: To further enhance robustness under occlusion and viewpoint variation, we employ Qwen-Image-Edit~\cite{wu2025qwenimagetechnicalreport} to synthesize additional viewpoints from the original image, implicitly capturing 3D scene layout through view consistency.

\begin{figure*}[t]
    \centering
    \includegraphics[width=1.0\linewidth]{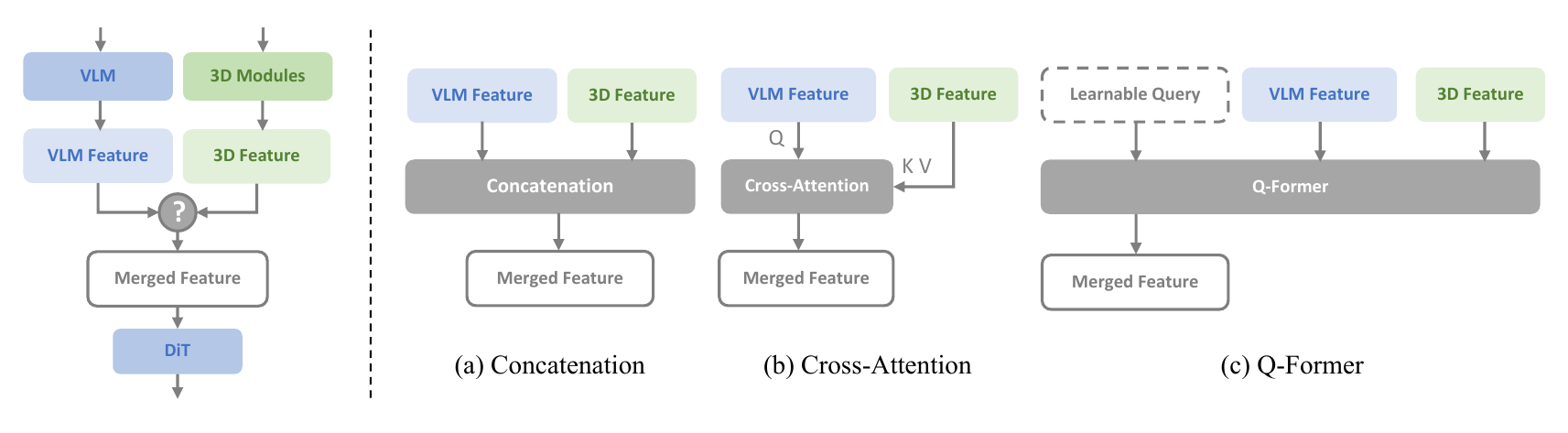}
    \caption{\textbf{3D Information Injection.}
    Left: pipeline for fusing VLM and 3D features. Right: three fusion strategies.
    }
    \label{fig:3d}
%\vspace{-1em}
\end{figure*}

Crucially, these 3D features are fused with the final-layer VLM features (identified in \Cref{subsec:vlm_feat}) as the most effective semantic representation. As illustrated in \Cref{fig:3d}, fusion occurs just before the action expert. We compare three different fusion strategies: concatenation, cross-attention (VLM features as queries, 3D features as keys/values), and Q-Former (learnable queries attending to multi-layer VLM and 3D features). Ablation studies in \Cref{sec:evaluation} show that single-layer cross-attention achieves optimal performance, confirming that attention-based interaction is key to harmonizing semantic and geometric streams.

For multi-view features, we lightly fine-tune Qwen-Image-Edit on Bridge~\cite{walke2023bridgedata} and collected LIBERO~\cite{liu2023libero} data, using only 50 paired samples per dataset. Experiments reveal that two synthesized views significantly outperform one, particularly on viewpoint-sensitive tasks (e.g., +14\% points on LIBERO-Plus’s camera perturbation subset). Our final system thus adopts two synthetic views with two-step inference, with their representations fused into the action expert via cross-attention to enrich spatial awareness.

Note the 3D module is fully modular. It can be toggled on/off or combined (single-view + multi-view) based on task demands, offering flexible deployment without retraining the core VLM. 
The semantic understanding from the VLM and geometric grounding from 3D models  together enables robust and precise action generation in complex 3D environments.

\section{Evaluation}
\label{sec:evaluation}
\subsection{Experiment Settings}
To comprehensively evaluate \VLAFM{}, we conducted extensive experiments on multiple simulation benchmarks, including LIBERO~\cite{liu2023libero}, LIBERO-Plus~\cite{fei25libero-plus}, RoboCasa GR1 Tabletop Tasks~\cite{GR00T} and Robotwin2.0~\cite{chen2025robotwin}, to assess the model's generalization, robustness, and adaptability to both single and dual-arm robots. Our training pipeline is built upon the StarVLA~\cite{starvla2025} framework. \VLAFM{} employs Qwen3-VL~\cite{Bai2025Qwen3VLTR} 4B as its VLM backbone, the 0.16B DiT with Action Manifold Learning (AML) as the action expert, and the VGGT~\cite{wang2025vggt} as an optional 3D spatial model. By default, we use 4 denoising steps and an action chunk size of 16. For visual inputs, images are resized to 224 $\times$ 224. For pre-training, we use a learning rate of 1e-5, a total batch size of 1024, and train the model for 100K steps.

\subsection{Main Results}

\subsubsection{LIBERO}
\begin{table}[h]
\centering
\caption{\textbf{Evaluation results on the LIBERO benchmark.} We train one \VLAFM{} jointly on all suites and report the success rate on each suite.}
\label{tab:libero_results}
\small
\setlength{\tabcolsep}{8pt}
\resizebox{.8\linewidth}{!}{
\begin{tabular}{lccccc}
\toprule
\textbf{Method} & \textbf{L-Spatial} & \textbf{L-Object} & \textbf{L-Goal} & \textbf{L-Long} & \textbf{Average} \\
\midrule
Diffusion Policy~\cite{Diffusionpolicy}  & 78.5 & 87.5 & 73.5 & 64.8 & 76.1 \\
OpenVLA~\cite{kim2024openvla}  & 84.7 & 88.4 & 79.2 & 53.7 & 76.5 \\
SpatialVLA~\cite{qu2025spatialvla}  & 88.2 & 89.9 & 78.6 & 55.5 & 78.1 \\
CoT-VLA~\cite{zhao2025cotvla}  & 87.5 & 91.6 & 87.6 & 69.0 & 83.9 \\
$\pi_0$-Fast~\cite{pertsch2025fast}  & 96.4 & 96.8 & 88.6 & 60.2 & 85.5 \\
GR00T-N1~\cite{GR00T}  & 94.4 & 97.6 & 93.0 & 90.6 & 93.9 \\
$\pi_0$~\cite{black2024pi_0}  & 98.0 & 96.8 & 94.4 & 88.4 & 94.4 \\
F1~\cite{lv2025f1}  & 98.2 & 97.8 & 95.4 & 91.3 & 95.7 \\
InternVLA-M1~\cite{internvlam1}  & 98.0 & 99.0 & 93.8 & 92.6 & 95.9 \\
Discrete Diffusion VLA~\cite{liang2025discrete}  & 97.2 & 98.6 & 97.4 & 92.0 & 96.3 \\
$\pi_{0.5}$~\cite{pi_0.5} & \textbf{98.8} & 98.2 & 98.0 & 92.4 & 96.9 \\
GR00T-N1.6~\cite{GR00T} & 97.7 &98.5&97.5&94.4& 97.0\\
OpenVLA-OFT~\cite{OFT}  & 97.6 & 98.4 & 97.9 & 94.5 & 97.1 \\
X-VLA~\cite{xvla} & 98.2 & 98.6 & 97.8 & \textbf{97.6} & 98.1 \\
% EO1~\cite{eo1}  & \textbf{99.7} & \textbf{99.8} & \underline{99.2} & 94.8 & \underline{98.2} \\
\midrule
\rowcolor{lightorange}
\textbf{\VLAFM{} (Ours)}  & \textbf{98.8} & \textbf{99.8} & \textbf{99.0} & 96.6 & \textbf{98.6} \\
\bottomrule
\end{tabular}
}
\label{tab:libero}
\end{table}
The main experimental results on the LIBERO~\cite{liu2023libero} benchmark are presented in \Cref{tab:libero}. \VLAFM{} demonstrates outstanding performance across the LIBERO test suites, achieving an average success rate of 98.6\%. Notably, it attained success rates of 98.8\% and 96.6\% on the spatial long-horizon trajectory tasks, respectively. These results indicate that \VLAFM{} possesses strong spatial understanding capabilities and can effectively execute complex multi-step manipulation tasks.

\subsubsection{LIBERO Plus}
\begin{table}[h]\centering
% \vspace{-10pt}
\caption{\textbf{Zero-shot performance on LIBERO-Plus.} All methods are trained only on the standard LIBERO dataset without fine-tuning on LIBERO-Plus dataset.
% One \VLAFM{} for all 4 suites is trained only on the standard LIBERO dataset (no LIBERO-Plus data fine-tuning).
}
\resizebox{\linewidth}{!}{%
% \small
\begin{tabular}{lcccccccc}\toprule
\textbf{Method}&\textbf{Camera} &\textbf{Robot} &\textbf{Language} &\textbf{Light} &\textbf{Background} &\textbf{Noise} &\textbf{Layout} &\textbf{Total} \\
\midrule
OpenVLA~\cite{kim2024openvla} &0.8 &3.5 &23.0 &8.1 &34.8 &15.2 &28.5 &15.6 \\
OpenVLA-OFT~\cite{OFT} &56.4 &31.9 &79.5 &88.7 &93.3 &75.8 &74.2 &69.6 \\
OpenVLA-OFT\_w~\cite{OFT} &10.4 &38.7 &70.5 &76.8 &\textbf{93.6} &49.9 &69.9 &55.8 \\
Openvla-OFT\_m~\cite{OFT} &55.6 &21.7 &81.0 &92.7 &91.0 &78.6 &68.7 &67.9 \\
NORA~\cite{hung2025nora} &2.2 &37.0 &65.1 &45.7 &58.6 &12.8 &62.1 &39.0 \\
WorldVLA~\cite{cen2025worldvla} &0.1 &27.9 &41.6 &43.7 &17.1 &10.9 &38.0 &25.0 \\
UniVLA~\cite{bu2025univla} &1.8 &46.2 &69.6 &69.0 &81.0 &21.2 &31.9 &42.9 \\
$\pi_0$~\cite{black2024pi_0} &13.8 &6.0 &58.8 &85.0 &81.4 &79.0 &68.9 &53.6 \\
$\pi_0$-Fast~\cite{pertsch2025fast} &\textbf{65.1} &21.6 &61.0 &73.2 &73.2 &74.4 &68.8 &61.6 \\
RIPT-VLA~\cite{RIPTVLA} &55.2 &31.2 &77.6 &88.4 &91.6 &73.5 &74.2 &68.4 \\
\midrule
\rowcolor{lightorange}
\textbf{\VLAFM{} (Ours)} &60.4&\textbf{67.9}&\textbf{86.4}&\textbf{96.2}&91.6&\textbf{86.4}&\textbf{82.6}&\textbf{80.5} \\
\bottomrule
\end{tabular}
}
% \vspace{-10pt}
\label{tab:libero-plus}
\end{table}
To further assess the generalization ability and robustness of \VLAFM{} in unseen scenarios, we also evaluated it on LIBERO-Plus~\cite{fei25libero-plus}. LIBERO-Plus is a variant of LIBERO that introduces controllable perturbations across seven dimensions, including vision and language, for systematic vulnerability analysis. As shown in \Cref{tab:libero-plus}, we conducted zero-shot evaluation on LIBERO-Plus using our \VLAFM{} model trained solely on LIBERO. \VLAFM{} achieves a success rate of 80.5\%, substantially outperforming prior methods like OpenVLA and OFT by 12.6-64.9\%. This result demonstrates the model's strong inherent robustness and generalization against visual and language perturbations.

\subsubsection{RoboCasa GR1 Tabletop Tasks}

\begin{table}[h]
\centering
\caption{\textbf{Evaluation Results on RoboCasa GR1 Tabletop Tasks.} We train one model jointly on all 24 tasks, and report mean results over 50 rollouts per task.}
\label{tab:robocasa-results}
\setlength{\tabcolsep}{1pt}
\begin{adjustbox}{max width=\textwidth}
\begin{tabular}{lccccc >{\columncolor{lightorange}}c}
\toprule
\textbf{Task} & GR00T-N1.6 & Qwen3GR00T & Qwen3PI & Qwen3OFT & Qwen3FAST & \textbf{\VLAFM{} (Ours)} \\

\midrule
PnPBottleToCabinetClose & 51.5 & 46.0 & 26.0 & 30.0 & 38.0 &\textbf{86.0}\\
PnPCanToDrawerClose & 13.0 & \textbf{80.0} & 62.0 & 76.0 & 44.0 &74.0\\
PnPCupToDrawerClose & 8.5 & 54.0 & 42.0 & 44.0 & \textbf{56.0} &48.0\\
PnPMilkToMicrowaveClose & 14.0 & 48.0 & \textbf{50.0} & 44.0 & 44.0 &46.0\\
PnPPotatoToMicrowaveClose & 41.5 & 28.0 & 42.0 & 32.0 & 14.0 &\textbf{50.0}\\
PnPWineToCabinetClose & 16.5 & 46.0 & 32.0 & 36.0 & 14.0 &\textbf{66.0}\\
\addlinespace
PnPNovelFromCuttingboardToBasket & 58.0 & 48.0 & 40.0 & 50.0 & 54.0 &\textbf{70.0}\\
PnPNovelFromCuttingboardToCardboardbox & 46.5 & 40.0 & 46.0 & 40.0 & 42.0 &\textbf{58.0}\\
PnPNovelFromCuttingboardToPan & 68.5 & 68.0 & 60.0 & 70.0 & 58.0 &\textbf{76.0}\\
PnPNovelFromCuttingboardToPot & 65.0 & 52.0 & 40.0 & 54.0 & 58.0 &\textbf{66.0}\\
PnPNovelFromCuttingboardToTieredbasket & 46.5 & \textbf{56.0} & 44.0 & 38.0 & 40.0 &38.0\\
\addlinespace
PnPNovelFromPlacematToBasket & \textbf{58.5} & 42.0 & 44.0 &32.0 & 36.0 &52.0\\
PnPNovelFromPlacematToBowl & 57.5 & 44.0 & 52.0 & 58.0 & 38.0 &\textbf{66.0}\\
PnPNovelFromPlacematToPlate & \textbf{63.0} & 48.0 & 50.0 & 52.0 & 42.0 &60.0\\
PnPNovelFromPlacematToTieredshelf & \textbf{28.5} & 18.0 & 28.0 & 24.0 & 18.0 &26.0\\
\addlinespace
PnPNovelFromPlateToBowl & 57.0 & \textbf{60.0} & 52.0 & \textbf{60.0} & 52.0 &54.0\\
PnPNovelFromPlateToCardboardbox & 43.5 & \textbf{50.0} & 40.0 & \textbf{50.0} & 30.0 &48.0\\
PnPNovelFromPlateToPan & 51.0 & 54.0 & 36.0 & \textbf{66.0} & 48.0 &\textbf{66.0}\\
PnPNovelFromPlateToPlate & \textbf{78.7} & 70.0 & 48.0 & 68.0 & 50.0 &64.0\\
\addlinespace
PnPNovelFromTrayToCardboardbox & 51.5 & 38.0 & 34.0 & 44.0 & 28.0 &\textbf{54.0}\\
PnPNovelFromTrayToPlate & \textbf{71.0} & 56.0 & 64.0 & 56.0 & 34.0 &68.0\\
PnPNovelFromTrayToPot & \textbf{64.5} & 50.0 & 44.0 & 62.0 & 46.0 &64.0\\
PnPNovelFromTrayToTieredbasket & 57.0 & 36.0 & 50.0 & 54.0 & 36.0 &\textbf{60.0}\\
PnPNovelFromTrayToTieredshelf & 31.5 & 16.0 & 28.0 & 30.0 & 16.0 &\textbf{38.0}\\
\midrule
\textbf{Average} & 47.6 & 47.8 & 43.9 & 48.8 & 39.0 &\textbf{58.3}\\
\bottomrule
\end{tabular}
\end{adjustbox}
\end{table}

We further evaluated our method on the Robocasa GR1 Tabletop Tasks~\cite{GR00T}. This benchmark comprises 24 distinct manipulation tasks involving complex interactions with articulated objects of various geometries, designed to assess general-purpose robot policies in a wide range of household scenarios. In this evaluation, we predict a 29 action space encompassing both dual arms, hands, and the waist, with an action chunk size of 16, to test the robustness of the new AML paradigm for predicting 464 high-dimensional actions under limited information capacity. As shown in \Cref{tab:robocasa-results}, \VLAFM{} achieves a 58.3\% success rate, comprehensively outperforming prior noise prediction-based experts like GR00T-N1.6 and action regression methods such as OFT, thereby setting a new SOTA. These results demonstrate that, under the Action Manifold Hypothesis, directly predicting actions is a more effective paradigm than predicting noise, validating the superiority of AML.

\subsubsection{RoboTwin 2.0}

\begin{table*}[h]
  \centering
  \footnotesize
  \setlength{\tabcolsep}{5.5pt}
  \caption{\textbf{Evaluation results on RoboTwin 2.0 Benchmark.} (Clean vs Randomized, 50+ tasks).}
  \begin{tabular}{
    *{1}{>{\centering\arraybackslash}m{4.2cm}} 
    *{4}{>{\centering\arraybackslash}m{0.98cm}} 
    *{2}{>{\columncolor{lightorange}\centering\arraybackslash}m{0.98cm}} %<- 修改在这里
  }
    \toprule
    \textbf{\multirow{2}{*}{Simulation Task}} 
      & \multicolumn{2}{c}{$\mathbf{\pi}_{\mathbf{0.5}}$}
      & \multicolumn{2}{c}{X-VLA}
      % 这里我们为表头也添加背景色，以保持一致性
      & \multicolumn{2}{c}{\cellcolor{lightorange}{\textbf{\VLAFM{} (Ours)}}} \\
    & \textbf{Clean} & \textbf{Rand.} 
    & \textbf{Clean} & \textbf{Rand.} 
    & \textbf{Clean} & \textbf{Rand.}  \\
    \midrule
    \textit{Place Dual Shoes} & 12  & 7  & 79  & \textbf{88}  & 80  & 80 \\
\textit{Move Stapler Pad} & 16  & 18  & \textbf{78}  & 73  & 57  & 61 \\
\textit{Stack Blocks Two} & 48  & 56  & 92  & 87  &96 &\textbf{98} \\
\textit{Scan Object} & 42  & 38  & 14  & 36  &85 &\textbf{86}  \\
\textit{Place Object Stand} & 74  & 65  & 86  & 88  &90 &\textbf{91}  \\
\textit{Place Fan} & 25  & 36  & 80  & 75  &\textbf{97} & 95 \\
\textit{Move Pillbottle Pad} & 33  & 29  & 73  & 71  &\textbf{94} & 86 \\
\textit{Pick Dual Bottles} & 10  & 6  & 47  & 36  &\textbf{70} &61  \\
\textit{Blocks Ranking Rgb} & 43  & 35  & 83  & 83  &\textbf{90} & 79 \\
\textit{......(50 tasks)} &-&-&-&-&-&- \\
\textit{Turn Switch} & 5  & 6  & 40  & 61  &55 &\textbf{66}  \\
\textit{Pick Diverse Bottles} & 5  & 3  & 58  & 36  &\textbf{71} &65  \\
\textit{Place Bread Basket} & 48  & 56  & 81  & 71  &\textbf{89} &86 \\
\textit{Stack Blocks Three} & 15  & 16  & 6  & 10  & \textbf{84}  &77 \\
\textit{Put Bottles Dustbin} & 12  & 9  & 74  & 77  &80 & \textbf{89} \\
\textit{Place Can Basket} & 19  & 25  & 49  & 52  &\textbf{72} &63 \\
\textit{Stamp Seal} & 36  & 23  & 76  & \textbf{82}  &72 &75 \\
\textit{Handover Block} & 18  & 19  & 73  & 37  &\textbf{72} &69  \\
\textit{Stack Bowls Three} & 33  & 35  & 76  &\textbf{86}  &80 &\textbf{86}  \\
\textit{Place Object Basket} & 43  & 36  & 44  & 39  &\textbf{91} &88  \\
\textit{Open Microwave} & 35  & 37  & 79  & 71   &\textbf{88} &84 \\
    \midrule
    \textbf{\textit{Average}} & 42.98 & 43.84 & 72.80 & 72.84 &\textbf{86.06} &85.08\\
    \bottomrule
  \end{tabular}
  \label{tab:robotwin-short}
\vspace{-0.3cm}
\end{table*}
To evaluate the generalization capability of our method, we conduct multi-task training on RoboTwin2.0~\cite{chen2025robotwin}: \VLAFM{} and all baselines are trained on 2500 demonstrations collected in clean scenes (50 per task) plus 25000 demonstrations gathered in heavily randomized scenes (500 per task). The randomizations include random backgrounds, table clutter, table height perturbations, and random lighting. The results for $\pi_{0.5}$ and X-VLA are taken from Motus~\cite{bi2025motus}. As shown in \Cref{tab:robotwin-short}, \VLAFM{}  outperforms strong VLA models like $\pi_{0.5}$  and X-VLA in both the clean and randomized multi-task settings on RoboTwin2.0, achieving a success rate of over 80\%.

\subsection{Ablation Study}

\subsubsection{Action Manifold Learning}
\begin{table}[h]\centering

\caption{\textbf{Ablation studies of Action Manifold Learning} on LIBERO-Plus benchmark.}
\setlength{\tabcolsep}{3pt}
\resizebox{\linewidth}{!}{

\begin{tabular}{lcccccccccc}\toprule
\textbf{Method}&\textbf{Hparams}&\textbf{Camera} &\textbf{Robot} &\textbf{Language} &\textbf{Light} &\textbf{Background} &\textbf{Noise} &\textbf{Layout} &\textbf{Total} \\
\midrule
Qwen3-VL-GR00T& Denoising Steps 4&41.0&61.2&86.5&91.3&91.6&53.8&73.4&69.3 \\
\VLAFM{}&Action Chunk 8 &39.9&63.7&85.9&90.8&93.1&60.6&76.8&71.0  \\
\midrule
Qwen3-VL-GR00T&\multirow{2}{*}{Denoising Steps 2} &35.4&59.7&86.7&90.2&90.1&49.8&73.3&67.2 \\
\VLAFM{}& &37.4&59.6&82.7&\textbf{93.1}&92.9&56.7&\textbf{80.5}&69.7  \\
\hdashline 
Qwen3-VL-GR00T&\multirow{2}{*}{Denoising Steps 10} &36.7&61.0&88.0&91.9&91.9&51.1&74.6&68.6 \\
\VLAFM{}& &\textbf{44.6}&56.9&81.4&90.2&90.1&62.9&78.3&70.2  \\
\midrule
Qwen3-VL-GR00T&\multirow{2}{*}{Action Chunk 10} &37.3&61.6&\textbf{88.7}&92.8&92.8&51.7&75.2&69.3 \\
\VLAFM{} &&42.5&\textbf{64.1}&86.3&92.7&\textbf{93.3}&\textbf{63.2}&78.0&\textbf{72.4}  \\
\hdashline 
\rowcolor{lightorange}
Qwen3-VL-GR00T&&7.9&31.7&64.9&83.2&73.6&24.0&55.2&45.7 \\
\rowcolor{lightorange}
$\Delta$ (Change) &                 & \textcolor{gray}{-33.1} & \textcolor{gray}{-29.5} & \textcolor{gray}{-21.6} & \textcolor{gray}{-8.1} & \textcolor{gray}{-18.0} & \textcolor{gray}{-29.8} & \textcolor{gray}{-18.2} & \textcolor{red}{-23.6} \\
\rowcolor{lightorange}
\VLAFM{}&&23.5&53.9&74.6&92.8&91.4&53.1&68.7&62.8   \\
\rowcolor{lightorange}
$\Delta$ (Change) &    \multirow{-4}{*}{Action Chunk 30}             & \textcolor{gray}{-16.4} & \textcolor{gray}{-9.8} & \textcolor{gray}{-11.3} & \textcolor{gray}{+2.0} & \textcolor{gray}{-1.7} & \textcolor{gray}{-7.5} & \textcolor{gray}{-8.1} & \textcolor{OliveGreen}{-8.2} \\
\bottomrule
\end{tabular}
}
% \vspace{-10pt}
\label{tab:aml_libero_plus}
\end{table}
To validate the superiority of AML, we present a comprehensive comparison in \Cref{tab:aml_libero_plus} between its action prediction paradigm and the noise prediction paradigm of GR00T. For a fair comparison, both GR00T and \VLAFM{} are initialized from the same Qwen3-VL-FAST pretrained model, with their action experts both sized at 0.16B. Furthermore, \VLAFM{} does not employ an additional 3D spatial module in this setup, ensuring that the only distinction between the two models is their prediction paradigm.

Under the default configuration of 4 denoising steps and an action chunk size of 8, \VLAFM{} outperforms GR00T by 1.7\%, indicating that under typical settings, directly predicting the action is more efficient than predicting noise. When varying only the number of denoising steps, \VLAFM{} consistently outperforms GR00T, especially in the extreme scenario of just 2 denoising steps. However, when the number of denoising steps is increased to 10, neither model shows performance gains, suggesting that both \VLAFM{} and GR00T can already achieve high-quality action generation efficiently.

When the action chunk size is increased to 10, GR00T's performance shows no significant change, whereas \VLAFM{}'s performance improves further to 72.4. This is because most tasks in LIBERO-Plus are simple, short-horizon tasks, for which a chunk size of 10 is a more optimal choice. Notably, as we increase the action chunk size further to 30, this oversized chunk becomes ill-suited for the majority of these simple, short-horizon tasks. Concurrently, a larger action chunk increases the dimensionality of the action sequence to be predicted, thereby increasing task difficulty. Under these conditions, GR00T's performance drops sharply by 23.6\%, while \VLAFM{} maintains an impressive 62.8\% success rate. This validates the "Action Manifold Hypothesis", which posits that predicting high-dimensional noise is a difficult and inefficient paradigm requiring vast information capacity, whereas directly predicting the action itself is a superior approach. Particularly for future challenges such as more difficult long-horizon tasks requiring longer action chunks, and dexterous manipulation or whole-body control demanding higher action dimensionality, the AML paradigm lays a solid foundation for the long-term advancement of embodied intelligence.

\subsubsection{VLM Feature Interaction}
\begin{table}[h]\centering
% \vspace{-10pt}
\caption{\textbf{Ablation studies of VLM feature interaction manners} on LIBERO-Plus benchmark.}
\setlength{\tabcolsep}{3pt}
\resizebox{\linewidth}{!}{%
% \small
\begin{tabular}{c|cc|ccccccccc}\toprule
\textbf{Layers}&\textbf{Feature}&\textbf{Query}&\textbf{Camera} &\textbf{Robot} &\textbf{Language} &\textbf{Light} &\textbf{Background} &\textbf{Noise} &\textbf{Layout} &\textbf{Total} \\
\midrule

\multirow{2}{*}{Last} & \checkmark  &$\times$&39.9&\textbf{63.7}&85.9&90.8&\textbf{93.1}&60.6&\textbf{76.8}&\textbf{71.0}  \\
 &$\times$ & \checkmark &38.4&61.5&87.2&91.6&90.3&59.1&75.3&70.0  \\
\midrule

\multirow{2}{*}{Intermediate} &  \checkmark &$\times$&38.2&49.9&88.7&92.7&89.5&\textbf{64.3}&73.6&69.0  \\
 &$\times$ & \checkmark &\textbf{42.3}&54.4&\textbf{89.1}&87.5&88.7&53.9&75.1&68.3  \\

\midrule

\multirow{3}{*}{Last 16 layers} & \checkmark  &$\times$&36.9&54.1&87.1&\textbf{93.9}&90.3&51.1&74.4&67.4  \\
 &$\times$ & \checkmark &25.8&57.5&80.9&88.7&87.5&37.3&74.1&65.2  \\
 & \checkmark  &\checkmark&25.0&50.4&88.5&90.1&88.5&50.0&70.3&63.8  \\

\bottomrule
\end{tabular}
}
% \vspace{-10pt}
\label{tab:vlm_libero_plus}
\end{table}
In \Cref{tab:vlm_libero_plus}, we investigate the overall effect of different feature conditions on action generation after the VLM has been pre-trained on robotics data. Conditioning on the features from the final hidden layer yields the highest success rate of 71\%, suggesting that this layer has become highly abstract and task-relevant, encapsulating the most critical information required for action generation. For the action queries, we employ 64 queries to strike a balance between efficiency and performance. Using queries alone results in performance slightly inferior to that of the original features. This indicates that robotics pre-training has already effectively aligned the VLM's internal feature space with the action space, rendering the benefit of an additional, from-scratch query interface marginal. In fact, it may even underperform the direct use of the already-optimized original features. Furthermore, concatenating features and queries as the conditioning input resulted in the lowest performance. We posit that this is because the pre-trained VLM has already formed a coherent internal representational structure across its layers. In such a case, injecting query features, which may introduce redundant or conflicting signals, can disrupt the learning process of the policy network. Future work could also explore assigning different weights to these two types of features or more effective ways to combine them. In summary, after robotics pre-training, the final layer features are well-adapted to the action domain and achieve strong performance, while the gains from incorporating intermediate layer features or additional action queries are limited.

\subsubsection{3D Information Injection}
\label{3d_sec}
To systematically validate the effectiveness of our proposed 3D feature injection mechanism, we conduct ablation studies on LIBERO~\cite{liu2023libero} and its more challenging extension, LIBERO-Plus~\cite{fei25libero-plus}. All experiments are initialized from the same pretrained vision-language model (Qwen3-VL 4B) and differ only in the components under evaluation, which are introduced via supervised fine-tuning (SFT). We strictly exclude other enhancements to ensure that any observed performance change can be unambiguously attributed to our 3D perception mechanism.

Results are reported in \Cref{tab:3d_libero} and \Cref{tab:3d_libero_plus}. First, we evaluate VGGT-based single-image 3D features under three fusion strategies: Concatenation, Cross-Attention, and Q-Former. In all cases, integrating VGGT features consistently improves performance on both benchmarks, confirming the general utility of single-view 3D priors for VLA tasks. Among the three, a single-layer Cross-Attention yields the best results and is adopted as the default fusion strategy in subsequent experiments.

We then assess implicit multi-view features generated by Qwen-Image-Edit~\cite{wu2025qwenimagetechnicalreport}. Incorporating synthesized views significantly enhances generalization, yielding absolute success rate gains of 2.8\% on LIBERO and 2-4\% on LIBERO-Plus. Notably, using two synthesized views further strengthens 3D spatial awareness. This improvement is most pronounced on the ``camera viewpoint perturbation'' subset, where accuracy increases by up to 14\%, demonstrating that multi-view information effectively mitigates reliance on fixed observation angles.

\begin{table}[h]
\centering
\caption{\textbf{Ablation studies of 3D feature injection} on LIBERO benchmark.}
\label{tab:3d_libero}
\small
\setlength{\tabcolsep}{8pt}
\resizebox{.8\linewidth}{!}{
\begin{tabular}{lccccc}
\toprule
\textbf{Method} & \textbf{L-Spatial} & \textbf{L-Object} & \textbf{L-Goal} & \textbf{L-Long} & \textbf{Average} \\
\midrule
Baseline &97.8&98.2&94.6&90.8&95.4\\
\midrule
VGGT (cross attention) &99.4&99.2&96.2&95.6&97.6\\
VGGT (concat)&99.4&98.6&96.2&93&96.8\\
VGGT (Q-former)&99.2&99.2&97.2&94&97.4\\
\midrule
Qwen-Image-Edit (1 view)&99.4&\textbf{99.4}&97.2&96.4&98.1\\
Qwen-Image-Edit (2 views)&\textbf{99.6}&99.0&\textbf{97.4}&\textbf{96.6}&\textbf{98.2}\\
\bottomrule
\end{tabular}
}
\end{table}
\begin{table}[!tp]\centering
% \vspace{-10pt}
\caption{\textbf{Ablation studies of 3D feature injection} on LIBERO-plus benchmark.}
\resizebox{\linewidth}{!}{%
% \small
\begin{tabular}{lccccccccc}\toprule
\textbf{Method}&\textbf{Camera} &\textbf{Robot} &\textbf{Language} &\textbf{Light} &\textbf{Background} &\textbf{Noise} &\textbf{Layout} &\textbf{Total} \\
\midrule
Baseline &32.9 &50.8 &86.3 &85.7 &86.3 &62.0&73.6 &66.4 \\
\midrule
VGGT (cross attention) &45.8&\textbf{53.8}&86.8&\textbf{97.2}&\textbf{93.9}&\textbf{65.5}&69.8&\textbf{71.1}\\
VGGT (concat)&41.2&51.2&87.0&93.7&88.4&63.9&70.6&68.9\\
VGGT (Q-former)&44.3&51.0&87.2&95.1&91.3&62.0&\textbf{71.0}&69.6\\
\midrule
Qwen-Image-Edit (1 view)&38.5&52.6&87.5&94.9&90.3&62.5&64.6&68.0\\
Qwen-Image-Edit (2 views)&\textbf{46.7}&53.5&\textbf{87.6}&96.6&91.8&60.6&69.5&70.2\\
\bottomrule
\end{tabular}
}
% \vspace{-10pt}
\label{tab:3d_libero_plus}
\end{table}

% \\
% \\
% \section{Limitations \& Conclusion}
\section{Future Work \& Conclusion}

\label{sec:conclusion}

% \subsection{Limitations}
% \VLAFM{} advances general-purpose embodied intelligence but has key limitations. 3D spatial understanding uses external modules like VGGT and Qwen-Image-Edit, trained separately and integrated without joint optimization with the visual language model (VLM). Feed-forward fusion via concatenation or simple attention often yields weak semantic alignment and inefficient feature interaction. Sensor generalization is limited to RGB images and proprioceptive joint states, with no tactile feedback, force-torque signals, or IMU data. Thus, the model struggles with high-contact tasks such as screwing or insertion under friction, restricting real-world industrial use. Data curation relies on hand-designed rules for instruction normalization, anomaly detection, and action consistency. While UniACT-dataset delivers a high-quality subset via systematic filtering, this rule-based approach does not scale to diverse, complex datasets and lacks automatic error correction.
% \VLAFM{} 推进了通用具身智能，但仍存在关键局限。三维空间理解依赖于独立训练的 VGGT 和 Qwen-Image-Edit 等外部模块，且未与视觉语言模型（VLM）联合优化，前馈融合方式（如拼接或简单注意力）常导致语义对齐弱和特征交互低效。传感器泛化能力仅限于 RGB 图像和本体感知关节状态，缺乏触觉反馈、力矩信号或 IMU 数据，因此在螺丝拧紧或摩擦条件下插拔等高接触力任务中表现不佳，限制了其在真实工业场景中的应用。数据清洗依赖人工设计的规则进行指令归一化、异常检测和动作一致性检查。尽管 UniACT-dataset 通过系统过滤提供了高质量子集，但该基于规则的方法难以扩展到更复杂多样的数据集，且缺乏自动纠错能力。

\subsection{Future Work}
We will further take scaling data with human demonstrations and UMI-collected~\cite{chi2024umi,zhaxizhuoma2025fastumi} trajectories to approach data capacity limits. Finer control over data quality, especially task balance and embodiment coverage, requires deeper study. A self-evolving data engine could form a closed loop of execution, failure analysis, augmentation, and model update, using model predictions to guide data collection and labeling, reducing manual cleaning and improving data-policy co-adaptation. Unifying multi-modal sensing—force, touch, temperature—into a single action space enables end-to-end perception-decision-action for active interaction. 3D representation learning can shift from post-hoc injection to intrinsic modeling, learning geometric priors during pre-training via self-supervised depth and pose estimation. True ``one-brain, many-forms'' intelligence demands generalization to legged systems, drones, and humanoids through architectures that abstract hardware details and learn from universal physical principles.
% 未来方向包括引入人类演示和 UMI 采集轨迹以扩大数据规模，逼近数据容量极限。对数据质量的精细控制——尤其是任务平衡性和本体覆盖——仍需深入研究。构建自演化数据引擎，形成执行、故障分析、数据增强与模型更新的闭环，利用模型预测指导数据采集与标注，可减少人工清洗，提升数据与策略的协同适应。将力、触觉、温度等多模态感知统一至单一动作空间，可实现端到端的感知-决策-行动框架，支持主动交互。三维表征学习可从后处理注入转向内生建模，在预训练中通过自监督深度与姿态估计算法学习几何先验。实现真正的“一脑多形”智能需推广至足式系统、无人机和人形机器人，依赖能抽象硬件细节并基于通用物理规律学习策略的架构。

\subsection{Conclusions}
We present \VLAFM{}, a unified framework that jointly optimizes data standardization and architecture to build hardware-agnostic embodied agents. Integrating six major open-source datasets, we build UniACT-dataset, a foundation of 6M+ trajectories, and apply delta actions in the end-effector frame, pad-to-dual-arm modeling, and dual-level reweighting to unify sources and support cross-embodiment transfer. We introduce the \textit{Action Manifold Hypothesis} and \textit{Action Manifold Learning (AML)}, shifting action prediction from denoising to direct generation, improving decoding efficiency and policy robustness. A plug-and-play dual-stream architecture fuses VLM semantics with enhanced 3D perception. Experiments show \VLAFM{} achieves state-of-the-art performance on Libero, Libero-Plus, RoboCasa and Robotwin, surpassing pi0 and UniVLA. These results confirms that high-performance embodied intelligence is achievable without proprietary data or custom hardware when public resources are systematically engineered. We open-source all pipelines and code to advance community-driven progress. This effort advocates a new paradigm that embodied intelligence evolving through open collaboration, growing smarter via shared knowledge.
% 本文提出 \VLAFM{}，一种协同优化数据标准化与架构设计的统一框架，用于构建不依赖特定硬件的具身智能体。通过整合六大开源数据集，构建了包含六百万以上轨迹的 UniACT-dataset 数据基底，采用末端执行器坐标系下的增量动作、pad-to-dual 建模与双层重加权，统一异构来源并支持跨本体知识迁移。提出“动作流形假说”与 AML 机制，将动作预测由去噪转为直接生成，提升解码效率与策略鲁棒性。设计即插即用双流架构，融合 VLM 语义理解与增强的三维感知。实验表明，\VLAFM{} 在 Libero、Libero-Plus 和 RoboCasa 上实现当前最优的零样本性能，优于 pi0 和 uniVLA。结果表明，无需私有数据或定制硬件，只要对公共资源进行系统化工程处理，即可构建高性能具身智能。我们开源全部数据流程与代码，推动社区共建。该工作倡导一种新范式：具身智能通过开放协作持续演化，依靠共享知识不断进步。

\clearpage
% \section{Contributions and Acknowledgments}
\section{Contributions}
\label{sec:contributions}
\setlength{\parskip}{0pt} % 让段落之间没有额外空隙
\setlength{\itemsep}{0pt} % 如果用itemize
\setlength{\parsep}{0pt}  % 控制段落间距
\raggedcolumns

% \subsubsection*{Contributions}
Author contributions in the following areas are as follows:

\begin{itemize}
    \item \textbf{Data Collection \& Analysis:} Yandan Yang, Haoyun Liu, Ronghan Chen, Yuzhi Chen, Dekang Qi
    \item \textbf{Data Standardization:} Yandan Yang, Tong lin, Xinyuan Chang
    \item \textbf{Data Pipeline:} Tong Lin, Shuang Zeng, Dongjie Huo
    \item \textbf{Model Architecture:} Shuang Zeng, Junjin Xiao
    \item \textbf{Training:} Shuang Zeng, Tong Lin, Junjin Xiao
    \item \textbf{Evaluation:} Junjin Xiao, Shuang Zeng, Tong Lin
    \item \textbf{Writing:} Yandan Yang, Xinyuan Chang, Shuang Zeng, Tong Lin, Dekang Qi, Junjin Xiao
    \item \textbf{Project Lead:} Xinyuan Chang, Feng Xiong
    \item \textbf{Advisor:}  Mu Xu$^\dagger$, Zhiheng Ma, Xing Wei
\end{itemize}

% \begin{multicols}{-2}
% \subsubsection*{Core Contributors}
%     \begin{itemize}

%         \item xxx$^\dagger$

%     \end{itemize}

%     \columnbreak % Force the remaining content to the second column
% % \cleanpage
%     \subsubsection*{Contributors}
%     \begin{itemize}
%         % contributor
%         \item xxx
%     \end{itemize}
% \end{multicols}

{\renewcommand{\thefootnote}{\fnsymbol{footnote}}\footnotetext[2]{Corresponding author: xumu.xm@alibaba-inc.com}}

% We thank xxx for their valuable discussions and assistance.

\clearpage

\bibliographystyle{plainnat}
\bibliography{main}

\begin{thebibliography}{52}
\providecommand{\natexlab}[1]{#1}
\providecommand{\url}[1]{\texttt{#1}}
\expandafter\ifx\csname urlstyle\endcsname\relax
  \providecommand{\doi}[1]{doi: #1}\else
  \providecommand{\doi}{doi: \begingroup \urlstyle{rm}\Url}\fi

\bibitem[Bai et~al.(2025)Bai, Cai, Chen, Chen, Chen, Cheng, Deng, Ding, Fang, Gao, Ge, Ge, Guo, Huang, Huang, Huang, Hui, Jiang, Li, Li, Li, Li, Lin, Lin, Liu, Liu, Liu, Liu, Liu, Liu, Lu, Luo, Lv, Men, Meng, Ren, yi~Ren, Song, Sun, Tang, Tu, Wan, Wang, Wang, Wang, Wang, Xie, Xu, Xu, Xu, Yang, Yang, Yang, Yang, Yu, Zhang, Zhang, Zhang, Zheng, Zhong, Zhou, Zhou, Zhou, Zhu, and Zhu]{Bai2025Qwen3VLTR}
Shuai Bai, Yuxuan Cai, Ruizhe Chen, Keqin Chen, Xiong-Hui Chen, Zesen Cheng, Lianghao Deng, Wei Ding, Rongyao Fang, Chang Gao, Chunjiang Ge, Wenbin Ge, Zhifang Guo, Qidong Huang, Qidong Huang, Fei Huang, Binyuan Hui, Shutong Jiang, Zhaohai Li, Mingsheng Li, Mei Li, Kaixin Li, Zicheng Lin, Junyang Lin, Xuejing Liu, Jiawei Liu, Chenglong Liu, Yang Liu, Dayiheng Liu, Shixuan Liu, Dunjie Lu, Ruilin Luo, Chenxu Lv, Rui Men, Li~Ying Meng, Xuancheng Ren, Xin yi~Ren, Sibo Song, Yu-Chen Sun, Jun Tang, Jianhong Tu, Jianqiang Wan, Peng Wang, Pengfei Wang, Qiuyue Wang, Yuxuan Wang, Tianbao Xie, Yihe Xu, Haiyang Xu, Jin Xu, Zhibo Yang, Mingkun Yang, Jianxin Yang, An~Yang, Bowen Yu, Fei Zhang, Hang Zhang, Xi~Zhang, Botao Zheng, Humen Zhong, Jingren Zhou, Fanxi Zhou, Jingren Zhou, Yuanzhi Zhu, and Keming Zhu.
\newblock Qwen3-vl technical report.
\newblock \emph{arXiv preprint arXiv:2511.21631}, 2025.

\bibitem[Bi et~al.(2025)Bi, Tan, Xie, Wang, Huang, Liu, Zhao, Feng, Xiang, Rong, et~al.]{bi2025motus}
Hongzhe Bi, Hengkai Tan, Shenghao Xie, Zeyuan Wang, Shuhe Huang, Haitian Liu, Ruowen Zhao, Yao Feng, Chendong Xiang, Yinze Rong, et~al.
\newblock Motus: A unified latent action world model.
\newblock \emph{arXiv preprint arXiv:2512.13030}, 2025.

\bibitem[Black et~al.(2024)Black, Brown, Driess, Esmail, Equi, Finn, Fusai, Groom, Hausman, Ichter, et~al.]{black2024pi_0}
Kevin Black, Noah Brown, Danny Driess, Adnan Esmail, Michael Equi, Chelsea Finn, Niccolo Fusai, Lachy Groom, Karol Hausman, Brian Ichter, et~al.
\newblock $pi\_0$: A vision-language-action flow model for general robot control.
\newblock \emph{arXiv preprint arXiv:2410.24164}, 2024.

\bibitem[Bu et~al.(2025)Bu, Yang, Cai, Gao, Ren, Yao, Luo, and Li]{bu2025univla}
Qingwen Bu, Yanting Yang, Jisong Cai, Shenyuan Gao, Guanghui Ren, Maoqing Yao, Ping Luo, and Hongyang Li.
\newblock Univla: Learning to act anywhere with task-centric latent actions.
\newblock \emph{RSS}, 2025.

\bibitem[Cadene et~al.(2024)Cadene, Alibert, Soare, Gallouedec, Zouitine, Palma, Kooijmans, Aractingi, Shukor, Aubakirova, Russi, Capuano, Pascal, Choghari, Moss, and Wolf]{cadene2024lerobot}
Remi Cadene, Simon Alibert, Alexander Soare, Quentin Gallouedec, Adil Zouitine, Steven Palma, Pepijn Kooijmans, Michel Aractingi, Mustafa Shukor, Dana Aubakirova, Martino Russi, Francesco Capuano, Caroline Pascal, Jade Choghari, Jess Moss, and Thomas Wolf.
\newblock Lerobot: State-of-the-art machine learning for real-world robotics in pytorch.
\newblock \url{https://github.com/huggingface/lerobot}, 2024.

\bibitem[Carlsson(2009)]{carlsson2009topology}
Gunnar Carlsson.
\newblock Topology and data.
\newblock \emph{Bulletin of the American Mathematical Society}, 46\penalty0 (2):\penalty0 255--308, 2009.

\bibitem[Cen et~al.(2025)Cen, Yu, Yuan, Jiang, Huang, Guo, Li, Song, Luo, Wang, et~al.]{cen2025worldvla}
Jun Cen, Chaohui Yu, Hangjie Yuan, Yuming Jiang, Siteng Huang, Jiayan Guo, Xin Li, Yibing Song, Hao Luo, Fan Wang, et~al.
\newblock Worldvla: Towards autoregressive action world model.
\newblock \emph{arXiv preprint arXiv:2506.21539}, 2025.

\bibitem[Chapelle et~al.(2006)Chapelle, Scholkopf, and Zien]{SSL}
Olivier Chapelle, Bernhard Scholkopf, and Alexander Zien.
\newblock \emph{Semi-Supervised Learning}.
\newblock The MIT Press, 09 2006.
\newblock ISBN 9780262033589.
\newblock \doi{10.7551/mitpress/9780262033589.001.0001}.
\newblock URL \url{https://doi.org/10.7551/mitpress/9780262033589.001.0001}.

\bibitem[Cheang et~al.(2025)Cheang, Chen, Cui, Hu, Huang, Kong, Li, Li, Liu, Ma, et~al.]{cheang2025gr}
Chilam Cheang, Sijin Chen, Zhongren Cui, Yingdong Hu, Liqun Huang, Tao Kong, Hang Li, Yifeng Li, Yuxiao Liu, Xiao Ma, et~al.
\newblock Gr-3 technical report.
\newblock \emph{arXiv preprint arXiv:2507.15493}, 2025.

\bibitem[Chen et~al.(2025)Chen, Chen, Chen, Cai, Liu, Li, Liang, Lin, Ge, Gu, et~al.]{chen2025robotwin}
Tianxing Chen, Zanxin Chen, Baijun Chen, Zijian Cai, Yibin Liu, Zixuan Li, Qiwei Liang, Xianliang Lin, Yiheng Ge, Zhenyu Gu, et~al.
\newblock Robotwin 2.0: A scalable data generator and benchmark with strong domain randomization for robust bimanual robotic manipulation.
\newblock \emph{arXiv preprint arXiv:2506.18088}, 2025.

\bibitem[Chi et~al.(2024)Chi, Xu, Pan, Cousineau, Burchfiel, Feng, Tedrake, and Song]{chi2024umi}
Cheng Chi, Zhenjia Xu, Chuer Pan, Eric Cousineau, Benjamin Burchfiel, Siyuan Feng, Russ Tedrake, and Shuran Song.
\newblock Universal manipulation interface: In-the-wild robot teaching without in-the-wild robots.
\newblock \emph{arXiv preprint arXiv:2402.10329}, 2024.

\bibitem[Chi et~al.(2025)Chi, Xu, Feng, Cousineau, Du, Burchfiel, Tedrake, and Song]{Diffusionpolicy}
Cheng Chi, Zhenjia Xu, Siyuan Feng, Eric Cousineau, Yilun Du, Benjamin Burchfiel, Russ Tedrake, and Shuran Song.
\newblock Diffusion policy: Visuomotor policy learning via action diffusion.
\newblock \emph{The International Journal of Robotics Research}, 44\penalty0 (10-11):\penalty0 1684--1704, 2025.

\bibitem[contributors(2024)]{agibotbeta}
AgiBot World~Colosseum contributors.
\newblock Agibot world colosseum.
\newblock \url{https://github.com/OpenDriveLab/AgiBot-World}, 2024.

\bibitem[Contributors(2025)]{internvlam1}
InternVLA-M1 Contributors.
\newblock Internvla-m1: A spatially guided vision-language-action framework for generalist robot policy.
\newblock \emph{arXiv preprint arXiv:2510.13778}, 2025.

\bibitem[Fei et~al.(2025)Fei, Wang, Shi, Dai, Cai, Qian, Ji, He, Zhang, Fei, Fu, Gong, and Qiu]{fei25libero-plus}
Senyu Fei, Siyin Wang, Junhao Shi, Zihao Dai, Jikun Cai, Pengfang Qian, Li~Ji, Xinzhe He, Shiduo Zhang, Zhaoye Fei, Jinlan Fu, Jingjing Gong, and Xipeng Qiu.
\newblock Libero-plus: In-depth robustness analysis of vision-language-action models.
\newblock \emph{arXiv preprint arXiv:2510.13626}, 2025.

\bibitem[Hung et~al.(2025)Hung, Sun, Hong, Zadeh, Li, Tan, Majumder, Poria, et~al.]{hung2025nora}
Chia-Yu Hung, Qi~Sun, Pengfei Hong, Amir Zadeh, Chuan Li, U~Tan, Navonil Majumder, Soujanya Poria, et~al.
\newblock Nora: A small open-sourced generalist vision language action model for embodied tasks.
\newblock \emph{arXiv preprint arXiv:2504.19854}, 2025.

\bibitem[Intelligence et~al.(2025)Intelligence, Black, Brown, Darpinian, Dhabalia, Driess, Esmail, Equi, Finn, Fusai, et~al.]{pi_0.5}
Physical Intelligence, Kevin Black, Noah Brown, James Darpinian, Karan Dhabalia, Danny Driess, Adnan Esmail, Michael Equi, Chelsea Finn, Niccolo Fusai, et~al.
\newblock $pi\_ $\{$0.5$\}$ $: a vision-language-action model with open-world generalization.
\newblock \emph{arXiv preprint arXiv:2504.16054}, 2025.

\bibitem[Ji et~al.(2025)Ji, Polavaram, Chen, Bajamahal, Ma, Adebola, Xu, and Goldberg]{ji2025oxeauge}
Guanhua Ji, Harsha Polavaram, Lawrence~Yunliang Chen, Sandeep Bajamahal, Zehan Ma, Simeon Adebola, Chenfeng Xu, and Ken Goldberg.
\newblock Oxe-auge: A large-scale robot augmentation of oxe for scaling cross-embodiment policy learning.
\newblock \emph{arXiv preprint arXiv:2512.13100}, 2025.

\bibitem[Jiang et~al.(2025)Jiang, Yuan, Liu, Lu, Cui, Liu, Cheng, Gao, Xu, and Zhao]{jiang2025galaxea}
Tao Jiang, Tianyuan Yuan, Yicheng Liu, Chenhao Lu, Jianning Cui, Xiao Liu, Shuiqi Cheng, Jiyang Gao, Huazhe Xu, and Hang Zhao.
\newblock Galaxea open-world dataset and g0 dual-system vla model.
\newblock \emph{arXiv preprint arXiv:2509.00576}, 2025.

\bibitem[Kim et~al.(2024)Kim, Pertsch, Karamcheti, Xiao, Balakrishna, Nair, Rafailov, Foster, Lam, Sanketi, et~al.]{kim2024openvla}
Moo~Jin Kim, Karl Pertsch, Siddharth Karamcheti, Ted Xiao, Ashwin Balakrishna, Suraj Nair, Rafael Rafailov, Ethan Foster, Grace Lam, Pannag Sanketi, et~al.
\newblock Openvla: An open-source vision-language-action model.
\newblock \emph{arXiv preprint arXiv:2406.09246}, 2024.

\bibitem[Kim et~al.(2025)Kim, Finn, and Liang]{OFT}
Moo~Jin Kim, Chelsea Finn, and Percy Liang.
\newblock Fine-tuning vision-language-action models: Optimizing speed and success.
\newblock \emph{RSS}, 2025.

\bibitem[Lee et~al.(2025)Lee, Duan, Fang, Deng, Liu, Li, Fang, Zhang, Wang, Lee, et~al.]{lee2025molmoact}
Jason Lee, Jiafei Duan, Haoquan Fang, Yuquan Deng, Shuo Liu, Boyang Li, Bohan Fang, Jieyu Zhang, Yi~Ru Wang, Sangho Lee, et~al.
\newblock Molmoact: Action reasoning models that can reason in space.
\newblock \emph{arXiv preprint arXiv:2508.07917}, 2025.

\bibitem[LejuRobotics(2025)]{LET2025}
LejuRobotics.
\newblock Let:full-size humanoid robot real-world dataset.
\newblock \url{https://huggingface.co/datasets/LejuRobotics/let_dataset}, 2025.

\bibitem[Li et~al.(2024)Li, Liang, Wang, Luo, Chen, Liao, Wei, Deng, Xu, Zhang, et~al.]{li2024cogact}
Qixiu Li, Yaobo Liang, Zeyu Wang, Lin Luo, Xi~Chen, Mozheng Liao, Fangyun Wei, Yu~Deng, Sicheng Xu, Yizhong Zhang, et~al.
\newblock Cogact: A foundational vision-language-action model for synergizing cognition and action in robotic manipulation.
\newblock \emph{arXiv preprint arXiv:2411.19650}, 2024.

\bibitem[Li and He(2025)]{JIT}
Tianhong Li and Kaiming He.
\newblock Back to basics: Let denoising generative models denoise.
\newblock \emph{arXiv preprint arXiv:2511.13720}, 2025.

\bibitem[Liang et~al.(2023)Liang, Bian, Xiao, Zhang, Chen, Liu, Xiang, Huang, and Su]{liang2023robo360}
Litian Liang, Liuyu Bian, Caiwei Xiao, Jialin Zhang, Linghao Chen, Isabella Liu, Fanbo Xiang, Zhiao Huang, and Hao Su.
\newblock Robo360: a 3d omnispective multi-material robotic manipulation dataset.
\newblock \emph{arXiv preprint arXiv:2312.06686}, 2023.

\bibitem[Liang et~al.(2025)Liang, Li, Yang, Wu, Mao, Nian, Pei, Zhou, Yang, Pang, et~al.]{liang2025discrete}
Zhixuan Liang, Yizhuo Li, Tianshuo Yang, Chengyue Wu, Sitong Mao, Tian Nian, Liuao Pei, Shunbo Zhou, Xiaokang Yang, Jiangmiao Pang, et~al.
\newblock Discrete diffusion vla: Bringing discrete diffusion to action decoding in vision-language-action policies.
\newblock \emph{arXiv preprint arXiv:2508.20072}, 2025.

\bibitem[Liu et~al.(2023)Liu, Zhu, Gao, Feng, Liu, Zhu, and Stone]{liu2023libero}
Bo~Liu, Yifeng Zhu, Chongkai Gao, Yihao Feng, Qiang Liu, Yuke Zhu, and Peter Stone.
\newblock Libero: Benchmarking knowledge transfer for lifelong robot learning.
\newblock \emph{arXiv preprint arXiv:2306.03310}, 2023.

\bibitem[Liu et~al.(2026)Liu, Li, Ma, Wu, Tan, Ouyang, Su, and Zhu]{liu2026rdt2}
Songming Liu, Bangguo Li, Kai Ma, Lingxuan Wu, Hengkai Tan, Xiao Ouyang, Hang Su, and Jun Zhu.
\newblock Rdt2: Exploring the scaling limit of umi data towards zero-shot cross-embodiment generalization.
\newblock \emph{arXiv preprint arXiv:2602.03310}, 2026.

\bibitem[Lv et~al.(2025)Lv, Kong, Li, Zeng, Qiu, Qu, Song, Chen, Deng, and Pang]{lv2025f1}
Qi~Lv, Weijie Kong, Hao Li, Jia Zeng, Zherui Qiu, Delin Qu, Haoming Song, Qizhi Chen, Xiang Deng, and Jiangmiao Pang.
\newblock F1: A vision-language-action model bridging understanding and generation to actions.
\newblock \emph{arXiv preprint arXiv:2509.06951}, 2025.

\bibitem[NVIDIA et~al.(2025)NVIDIA, Bjorck, Castañeda, Cherniadev, Da, Ding, Fan, Fang, Fox, Hu, Huang, Jang, Jiang, Kautz, Kundalia, Lao, Li, Lin, Lin, Liu, Llontop, Magne, Mandlekar, Narayan, Nasiriany, Reed, Tan, Wang, Wang, Wang, Wang, Xiang, Xie, Xu, Xu, Ye, Yu, Zhang, Zhang, Zhao, Zheng, and Zhu]{GR00T}
NVIDIA, Johan Bjorck, Fernando Castañeda, Nikita Cherniadev, Xingye Da, Runyu Ding, Linxi~"Jim" Fan, Yu~Fang, Dieter Fox, Fengyuan Hu, Spencer Huang, Joel Jang, Zhenyu Jiang, Jan Kautz, Kaushil Kundalia, Lawrence Lao, Zhiqi Li, Zongyu Lin, Kevin Lin, Guilin Liu, Edith Llontop, Loic Magne, Ajay Mandlekar, Avnish Narayan, Soroush Nasiriany, Scott Reed, You~Liang Tan, Guanzhi Wang, Zu~Wang, Jing Wang, Qi~Wang, Jiannan Xiang, Yuqi Xie, Yinzhen Xu, Zhenjia Xu, Seonghyeon Ye, Zhiding Yu, Ao~Zhang, Hao Zhang, Yizhou Zhao, Ruijie Zheng, and Yuke Zhu.
\newblock Gr00t n1: An open foundation model for generalist humanoid robots, 2025.
\newblock URL \url{https://arxiv.org/abs/2503.14734}.

\bibitem[O’Neill et~al.(2024)O’Neill, Rehman, Maddukuri, Gupta, Padalkar, Lee, Pooley, Gupta, Mandlekar, Jain, et~al.]{o2024oxe}
Abby O’Neill, Abdul Rehman, Abhiram Maddukuri, Abhishek Gupta, Abhishek Padalkar, Abraham Lee, Acorn Pooley, Agrim Gupta, Ajay Mandlekar, Ajinkya Jain, et~al.
\newblock Open x-embodiment: Robotic learning datasets and rt-x models: Open x-embodiment collaboration 0.
\newblock In \emph{2024 IEEE International Conference on Robotics and Automation (ICRA)}, pages 6892--6903. IEEE, 2024.

\bibitem[Peebles and Xie(2022)]{Peebles2022DiT}
William Peebles and Saining Xie.
\newblock Scalable diffusion models with transformers.
\newblock \emph{arXiv preprint arXiv:2212.09748}, 2022.

\bibitem[Pertsch et~al.(2025)Pertsch, Stachowicz, Ichter, Driess, Nair, Vuong, Mees, Finn, and Levine]{pertsch2025fast}
Karl Pertsch, Kyle Stachowicz, Brian Ichter, Danny Driess, Suraj Nair, Quan Vuong, Oier Mees, Chelsea Finn, and Sergey Levine.
\newblock Fast: Efficient action tokenization for vision-language-action models.
\newblock \emph{arXiv preprint arXiv:2501.09747}, 2025.

\bibitem[Qu et~al.(2025)Qu, Song, Chen, Yao, Ye, Ding, Wang, Gu, Zhao, Wang, et~al.]{qu2025spatialvla}
Delin Qu, Haoming Song, Qizhi Chen, Yuanqi Yao, Xinyi Ye, Yan Ding, Zhigang Wang, JiaYuan Gu, Bin Zhao, Dong Wang, et~al.
\newblock Spatialvla: Exploring spatial representations for visual-language-action model.
\newblock \emph{RSS}, 2025.

\bibitem[Ramos et~al.(2021)Ramos, Girgin, Hussenot, Vincent, Yakubovich, Toyama, Gergely, Stanczyk, Marinier, Harmsen, et~al.]{ramos2021rlds}
Sabela Ramos, Sertan Girgin, L{\'e}onard Hussenot, Damien Vincent, Hanna Yakubovich, Daniel Toyama, Anita Gergely, Piotr Stanczyk, Raphael Marinier, Jeremiah Harmsen, et~al.
\newblock Rlds: an ecosystem to generate, share and use datasets in reinforcement learning.
\newblock \emph{arXiv preprint arXiv:2111.02767}, 2021.

\bibitem[Reuss et~al.(2025)Reuss, Zhou, R{\"u}hle, Ya{\u{g}}murlu, Otto, and Lioutikov]{reuss2025flower}
Moritz Reuss, Hongyi Zhou, Marcel R{\"u}hle, {\"O}mer~Erdin{\c{c}} Ya{\u{g}}murlu, Fabian Otto, and Rudolf Lioutikov.
\newblock Flower: Democratizing generalist robot policies with efficient vision-language-action flow policies.
\newblock \emph{arXiv preprint arXiv:2509.04996}, 2025.

\bibitem[starVLA Contributors(2025)]{starvla2025}
starVLA Contributors.
\newblock Starvla: A lego-like codebase for vision-language-action model developing.
\newblock GitHub repository, 1 2025.
\newblock URL \url{https://github.com/starVLA/starVLA}.

\bibitem[Tan et~al.(2025)Tan, Dou, Zhao, and Kr{\"a}henb{\"u}hl]{RIPTVLA}
Shuhan Tan, Kairan Dou, Yue Zhao, and Philipp Kr{\"a}henb{\"u}hl.
\newblock Interactive post-training for vision-language-action models.
\newblock \emph{arXiv preprint arXiv:2505.17016}, 2025.

\bibitem[Team et~al.(2024)Team, Ghosh, Walke, Pertsch, Black, Mees, Dasari, Hejna, Kreiman, Xu, et~al.]{team2024octo}
Octo~Model Team, Dibya Ghosh, Homer Walke, Karl Pertsch, Kevin Black, Oier Mees, Sudeep Dasari, Joey Hejna, Tobias Kreiman, Charles Xu, et~al.
\newblock Octo: An open-source generalist robot policy.
\newblock \emph{arXiv preprint arXiv:2405.12213}, 2024.

\bibitem[Vincent et~al.(2010)Vincent, Larochelle, Lajoie, Bengio, Manzagol, and Bottou]{vincent2010stacked}
Pascal Vincent, Hugo Larochelle, Isabelle Lajoie, Yoshua Bengio, Pierre-Antoine Manzagol, and L{\'e}on Bottou.
\newblock Stacked denoising autoencoders: Learning useful representations in a deep network with a local denoising criterion.
\newblock \emph{Journal of machine learning research}, 11\penalty0 (12), 2010.

\bibitem[Walke et~al.(2023)Walke, Black, Zhao, Vuong, Zheng, Hansen-Estruch, He, Myers, Kim, Du, et~al.]{walke2023bridgedata}
Homer~Rich Walke, Kevin Black, Tony~Z Zhao, Quan Vuong, Chongyi Zheng, Philippe Hansen-Estruch, Andre~Wang He, Vivek Myers, Moo~Jin Kim, Max Du, et~al.
\newblock Bridgedata v2: A dataset for robot learning at scale.
\newblock In \emph{Conference on Robot Learning}, pages 1723--1736, 2023.

\bibitem[Wang et~al.(2025)Wang, Chen, Karaev, Vedaldi, Rupprecht, and Novotny]{wang2025vggt}
Jianyuan Wang, Minghao Chen, Nikita Karaev, Andrea Vedaldi, Christian Rupprecht, and David Novotny.
\newblock Vggt: Visual geometry grounded transformer.
\newblock In \emph{Proceedings of the IEEE/CVF Conference on Computer Vision and Pattern Recognition}, 2025.

\bibitem[Wang et~al.(2026)Wang, Ding, Li, Cui, Ge, Tong, Song, Zhao, Zhao, Hou, Huang, Tang, Wang, Zhang, Liu, and Wang]{wang2025vlaadapter}
Yihao Wang, Pengxiang Ding, Lingxiao Li, Can Cui, Zirui Ge, Xinyang Tong, Wenxuan Song, Han Zhao, Wei Zhao, Pengxu Hou, Siteng Huang, Yifan Tang, Wenhui Wang, Ru~Zhang, Jianyi Liu, and Donglin Wang.
\newblock Vla-adapter: An effective paradigm for tiny-scale vision-language-action model.
\newblock \emph{AAAI}, 2026.

\bibitem[Wu et~al.(2025{\natexlab{a}})Wu, Li, Zhou, Lin, Gao, Yan, ming Yin, Bai, Xu, Chen, Chen, Tang, Zhang, Wang, Yang, Yu, Cheng, Liu, Li, Zhang, Meng, Wei, Ni, Chen, Cao, Peng, Qu, Wu, Wang, Yu, Wen, Feng, Xu, Wang, Zhang, Zhu, Wu, Cai, and Liu]{wu2025qwenimagetechnicalreport}
Chenfei Wu, Jiahao Li, Jingren Zhou, Junyang Lin, Kaiyuan Gao, Kun Yan, Sheng ming Yin, Shuai Bai, Xiao Xu, Yilei Chen, Yuxiang Chen, Zecheng Tang, Zekai Zhang, Zhengyi Wang, An~Yang, Bowen Yu, Chen Cheng, Dayiheng Liu, Deqing Li, Hang Zhang, Hao Meng, Hu~Wei, Jingyuan Ni, Kai Chen, Kuan Cao, Liang Peng, Lin Qu, Minggang Wu, Peng Wang, Shuting Yu, Tingkun Wen, Wensen Feng, Xiaoxiao Xu, Yi~Wang, Yichang Zhang, Yongqiang Zhu, Yujia Wu, Yuxuan Cai, and Zenan Liu.
\newblock Qwen-image technical report, 2025{\natexlab{a}}.
\newblock URL \url{https://arxiv.org/abs/2508.02324}.

\bibitem[Wu et~al.(2024)Wu, Hou, Liu, Che, Ju, Yang, Li, Zhao, Xu, Yang, et~al.]{wu2024robomind}
Kun Wu, Chengkai Hou, Jiaming Liu, Zhengping Che, Xiaozhu Ju, Zhuqin Yang, Meng Li, Yinuo Zhao, Zhiyuan Xu, Guang Yang, et~al.
\newblock Robomind: Benchmark on multi-embodiment intelligence normative data for robot manipulation.
\newblock \emph{arXiv preprint arXiv:2412.13877}, 2024.

\bibitem[Wu et~al.(2025{\natexlab{b}})Wu, Liu, Xie, Wang, Li, Yang, Li, Zhu, Wu, Liu, et~al.]{wu2025robocoin}
Shihan Wu, Xuecheng Liu, Shaoxuan Xie, Pengwei Wang, Xinghang Li, Bowen Yang, Zhe Li, Kai Zhu, Hongyu Wu, Yiheng Liu, et~al.
\newblock Robocoin: An open-sourced bimanual robotic data collection for integrated manipulation.
\newblock \emph{arXiv preprint arXiv:2511.17441}, 2025{\natexlab{b}}.

\bibitem[Zeng et~al.(2025)Zeng, Chang, Xie, Liu, Bai, Pan, Xu, Wei, and Guo]{zheng25}
Shuang Zeng, Xinyuan Chang, Mengwei Xie, Xinran Liu, Yifan Bai, Zheng Pan, Mu~Xu, Xing Wei, and Ning Guo.
\newblock Futuresightdrive: Thinking visually with spatio-temporal cot for autonomous driving.
\newblock \emph{arXiv preprint arXiv:2505.17685}, 2025.

\bibitem[Zhao et~al.(2025)Zhao, Lu, Kim, Fu, Zhang, Wu, Li, Ma, Han, Finn, et~al.]{zhao2025cotvla}
Qingqing Zhao, Yao Lu, Moo~Jin Kim, Zipeng Fu, Zhuoyang Zhang, Yecheng Wu, Zhaoshuo Li, Qianli Ma, Song Han, Chelsea Finn, et~al.
\newblock Cot-vla: Visual chain-of-thought reasoning for vision-language-action models.
\newblock In \emph{CVPR}, 2025.

\bibitem[Zhaxizhuoma et~al.(2025)Zhaxizhuoma, Liu, Guan, Jia, Wu, Liu, Wang, Liang, Chen, Zhang, et~al.]{zhaxizhuoma2025fastumi}
Zhaxizhuom Zhaxizhuoma, Kehui Liu, Chuyue Guan, Zhongjie Jia, Ziniu Wu, Xin Liu, Tianyu Wang, Shuai Liang, Pengan Chen, Pingrui Zhang, et~al.
\newblock Fastumi: A scalable and hardware-independent universal manipulation interface with dataset.
\newblock In \emph{Conference on Robot Learning}, pages 3069--3093. PMLR, 2025.

\bibitem[Zheng et~al.(2025)Zheng, Li, Wang, Liu, Kang, Feng, Zheng, Zou, Chen, Zeng, et~al.]{xvla}
Jinliang Zheng, Jianxiong Li, Zhihao Wang, Dongxiu Liu, Xirui Kang, Yuchun Feng, Yinan Zheng, Jiayin Zou, Yilun Chen, Jia Zeng, et~al.
\newblock X-vla: Soft-prompted transformer as scalable cross-embodiment vision-language-action model.
\newblock \emph{ICLR}, 2025.

\bibitem[Zitkovich et~al.(2023)Zitkovich, Yu, Xu, Xu, Xiao, Xia, Wu, Wohlhart, Welker, Wahid, et~al.]{rt2}
Brianna Zitkovich, Tianhe Yu, Sichun Xu, Peng Xu, Ted Xiao, Fei Xia, Jialin Wu, Paul Wohlhart, Stefan Welker, Ayzaan Wahid, et~al.
\newblock Rt-2: Vision-language-action models transfer web knowledge to robotic control.
\newblock In \emph{Conference on Robot Learning}, pages 2165--2183. PMLR, 2023.

\end{thebibliography}

\clearpage
% \input{sections/contribution}
% \clearpage
\beginappendix

\section{Appendix}

\let\clearpage\relax

\end{document}